\documentclass{article}
\usepackage[preprint]{corl_2025} 

\usepackage{multicol}
\usepackage{amsmath}    
\usepackage{amssymb}    
\usepackage{bm}         
\usepackage{graphicx}   
\usepackage{natbib}
\usepackage[table]{xcolor}
\usepackage{amsthm}
\usepackage{amsfonts}
\usepackage{booktabs}
\usepackage{tikz}
\usetikzlibrary{calc}
\usepackage{subcaption}
\usepackage{wrapfig}
\usepackage{siunitx}

\usepackage[font=small]{caption}
\usepackage{enumitem}
\setlist[itemize,enumerate]{nosep,leftmargin=*}   

\usepackage{circledsteps}
\newcommand\myCircled[2][]{
\hspace{-0.2em}\Circled[fill color=black,inner color=white, #1]{\sffamily#2}\hspace{-0.2em}
}

\theoremstyle{plain} 
\newtheorem{definition}{Definition}[section]

\newtheorem{proposition}[definition]{Proposition}

\usepackage{titlesec}
\titlespacing*{\section}
  {0pt}{*0.2}{*0.1}  

\titlespacing*{\paragraph}
  {0pt}{*0.3}{*0.3}  

\newcommand{\headline}[1]{\noindent\textbf{#1}}
\newcommand{\subheadline}[1]{\noindent\textit{#1}}
\newcommand{\videourl}{\url{https://www.youtube.com/watch?v=PQTa5rV8d4g}}
\makeatletter
\def\blfootnote{\gdef\@thefnmark{}\@footnotetext}
\makeatother

\title{ReCoDe: Reinforcement Learning-based Dynamic Constraint Design for Multi-Agent Coordination}
\author{Michael Amir, Guang Yang, Zhan Gao, Keisuke Okumura,  Heedo Woo, Amanda Prorok
}
\newcommand{\authorsinfo}{
All authors are with the University of Cambridge.
KO is also with National Institute of Advanced Industrial Science and Technology (AIST).
Email: \{ma2151,gy268,zg292,,ko393,hw527, asp45\}@cst.cam.ac.uk.
This work was supported by ERC Project 949940 (gAIa).}

\begin{document}

\maketitle
\blfootnote{\authorsinfo}
\blfootnote{\texttt{Supplementary video: \videourl}}

\vspace{-5mm}
{
\begin{figure*}[h!]
\centering

\begin{minipage}{0.39\textwidth}
    \centering
    \includegraphics[height=5.2cm]{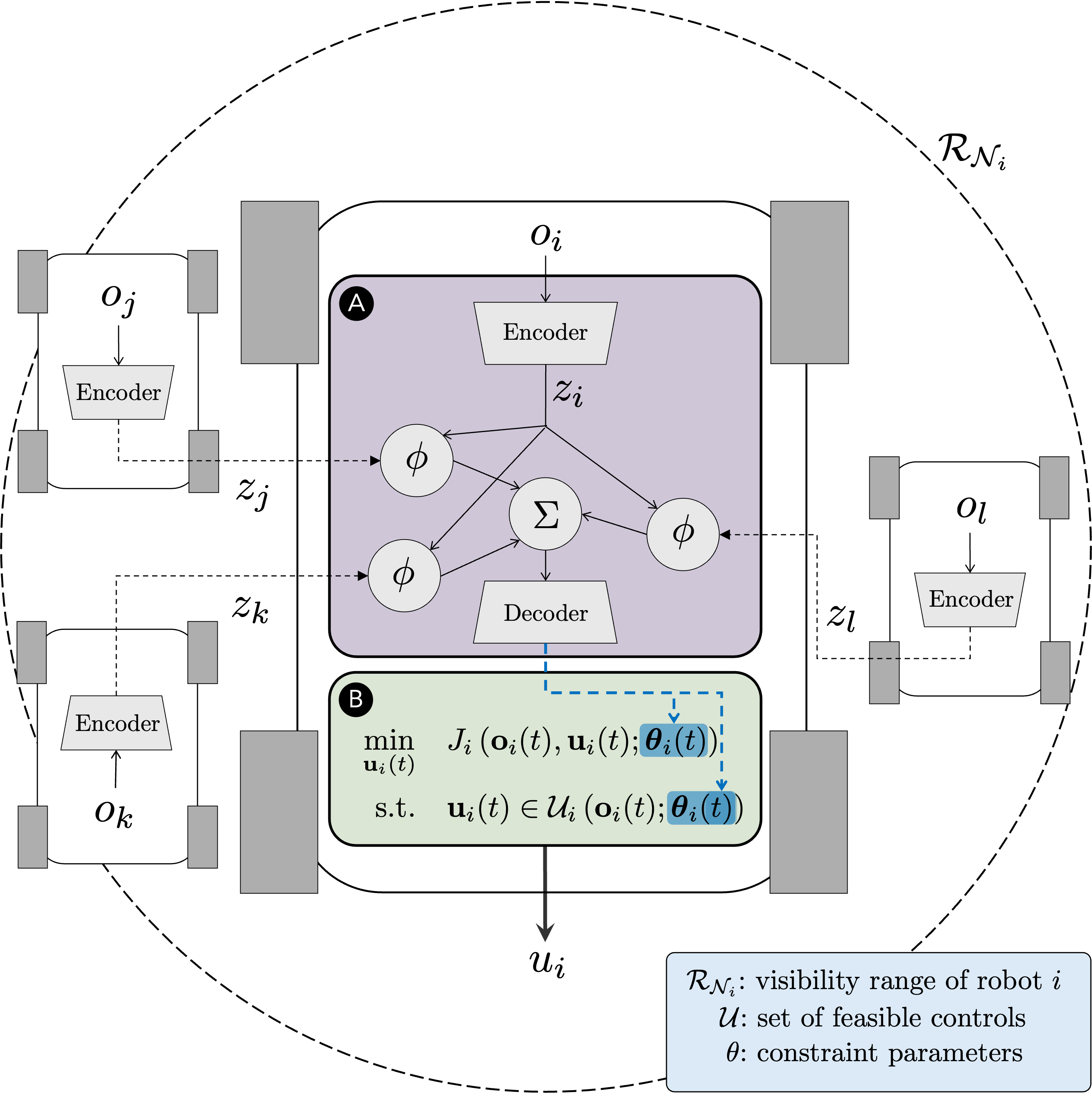}
\end{minipage}
\hfill
\begin{minipage}{0.6\textwidth}
    \centering
    \includegraphics[height=5.2cm]{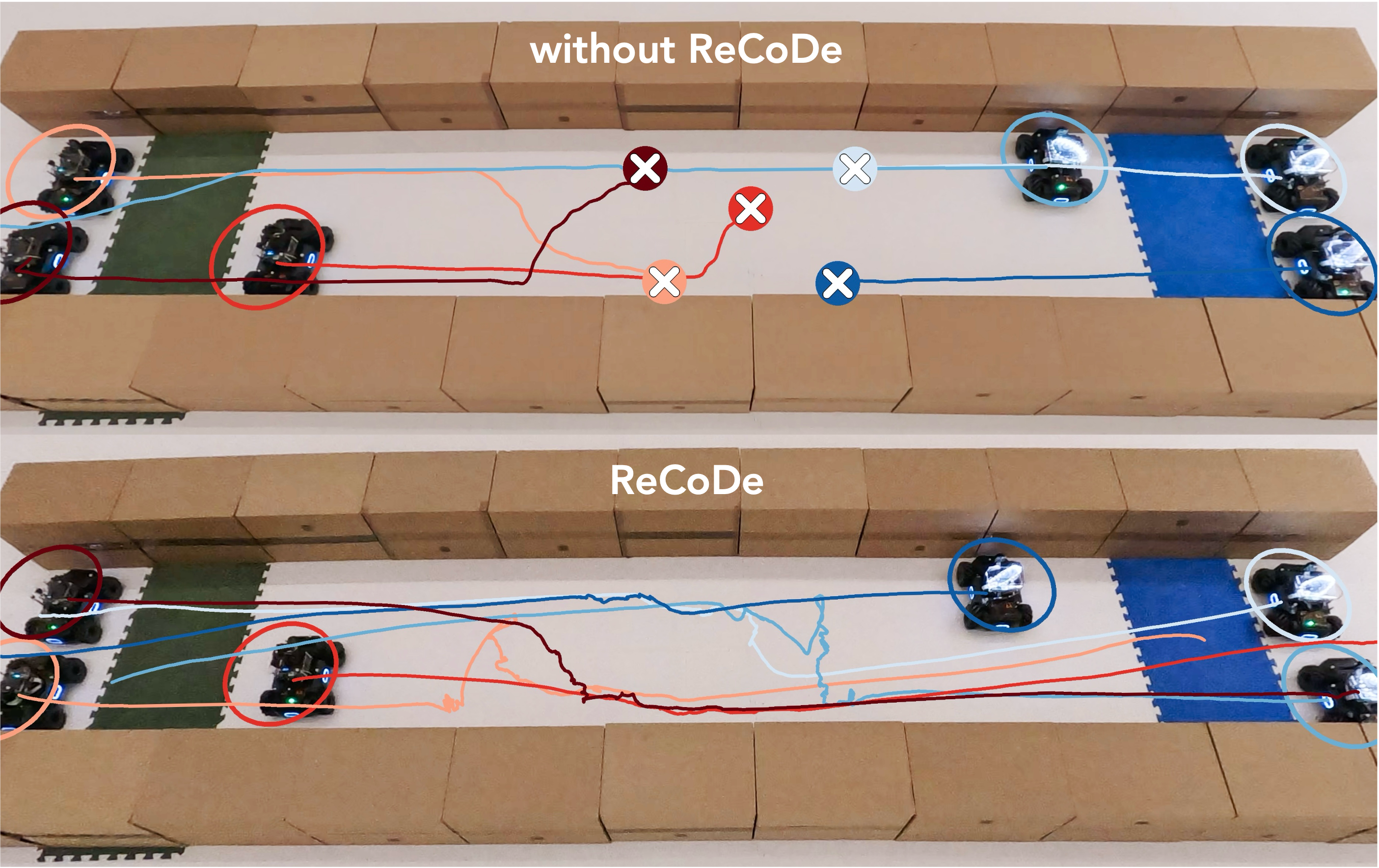}
\end{minipage}

\caption{
\textbf{Left:}
An overview of the proposed ReCoDe method. A GNN policy {\scriptsize\myCircled{A}} aggregates the encoded observations of neighboring agents within a visibility range $\mathcal{R}_{\mathcal{N}}$ and generates constraint parameters $\theta$ that influence the feasible set $\mathcal{U}$ of an optimization-based controller {\scriptsize\myCircled{B}}.
\textbf{Right}: Real-robot position-swap in a $90\,$cm-wide, $6.4\,$m-long corridor.  
\textbf{Top:} baseline QP controller dead-locks.  
\textbf{Bottom:} the \emph{same} controller augmented with ReCoDe succeeds.  
Six holonomic ground robots (\citeauthor{JanCam2024}) communicate only within $1.5\,$m.  
}
\label{fig:robomaster_exp}
\end{figure*}
}

\begin{abstract}
Constraint-based optimization is a cornerstone of robotics, enabling the design of controllers that reliably encode task and safety requirements such as collision avoidance or formation adherence. However, handcrafted constraints can fail in multi-agent settings that demand complex coordination. We introduce ReCoDe—Reinforcement-based Constraint Design—a decentralized, hybrid framework that merges the reliability of optimization-based controllers with the adaptability of multi-agent reinforcement learning. Rather than discarding expert controllers, ReCoDe improves them by learning additional, dynamic constraints that capture subtler behaviors, for example, by constraining agent movements to prevent congestion in cluttered scenarios. Through local communication, agents collectively constrain their allowed actions to coordinate more effectively under changing conditions. In this work, we focus on applications of ReCoDe to multi-agent navigation tasks requiring intricate, context-based movements and consensus, where we show that it outperforms purely handcrafted controllers, other hybrid approaches, and standard MARL baselines. We give empirical (real robot) and theoretical evidence that retaining a user-defined controller, even when it is imperfect, is more efficient than learning from scratch, especially because ReCoDe can dynamically change the degree to which it relies on this controller.


\end{abstract}

\section{Introduction}
\vspace{-1mm}
Ensuring that multiple autonomous agents, such as fleets of autonomous vehicles and warehouse robots,  can safely and efficiently coordinate in a shared environment is a long-standing challenge in robotics  \citep{alonso2015collision, merkt2019continuous}. Classical approaches rely heavily on optimization-based controllers, which encode mission objectives and constraints into a tractable optimization problem \citep{gregory2018constrained}. By carefully designing these constraints, we can ensure collision avoidance, enforce kinematic limitations, and guide agents toward their goals. Yet, as scenarios grow more complex, handcrafted constraints become insufficient: they cannot easily adapt to evolving conditions or fully exploit coordination opportunities \citep{gao2023onlinecbf}. At the opposite extreme, multi-agent reinforcement learning (MARL) is a highly adaptive paradigm that shapes agent behavior through experience without task-specific design \citep{busoniu2008comprehensive, canese2021multi, gronauer2022multi}. However, MARL lacks the analytic structure and safety assurances of optimization-based methods, making its decisions harder to predict, verify, or trust in critical applications.

This paper presents ReCoDe (Reinforcement-based Constraint Design), a hybrid, decentralized framework combining optimization-based control with adaptation provided by MARL. ReCoDe operates by \textit{augmenting} a given expert controller\footnote{We will refer to this controller, interchangeably, as the default/expert/handcrafted controller.}: it keeps the original controller (and its safety constraints) but learns additional, situation-dependent constraints via MARL. These learned constraints dynamically modify each agent's feasible action set, allowing for finer control and improved coordination beyond what the fixed controller permits. This approach preserves the desirable properties of the expert controller—like safety guarantees and interpretability---while enabling adaptation to complex scenarios. ReCoDe is inherently multi-agent, as agents learn to collectively shape their constraints for better overall performance. Agents integrate information from neighbors via local communication when deciding how to adjust their constraints at each point in time. This design is facilitated by a Graph Neural Network (GNN)-based policy (Figure \ref{fig:robomaster_exp}).


\subheadline{Why learn additional constraints?} While handcrafted optimization-based controllers excel at encoding single-agent safety requirements and tasks, they often fall short when dealing with more intricate multi-agent interactions. Consider a two-way street where vehicles want to travel either up or down: although handcrafted constraints can ensure safety (e.g., avoiding collisions), they typically do not address higher-level challenges such as congestion or mutual blocking. To deal with such challenges, human drivers impose on themselves additional  \textit{situational} constraints—like staying strictly within a lane or yielding to unblock traffic. These more nuanced behaviors hinge on communication among agents and online adaptation to changing conditions. Starting with only basic constraints (e.g., collision avoidance), ReCoDe learns such higher-level coordination rules as additional constraints, based on context and communication with other agents. The result is a hybrid controller that keeps formal guarantees, anchors learning with expert structure, and still adapts online to multi-agent interactions by letting agents collectively shape stricter constraints that help them coordinate.


\subheadline{Why learn constraints rather than the optimization objective?} Some existing hybrid approaches work by learning the optimization objective \cite{alshiekh2018safeshielding,romero2024actorcriticmodelpredictivecontrol}. We posit that retaining the handcrafted objective and refining decision-making with learned constraints is more data-efficient, because the default objective, even if imperfect, embeds expert knowledge of the task.  Moreover, while hard constraints can, in principle, be mimicked by infinite-cost penalties in the objective, thus learning the objective is \textit{theoretically} more general--RL rarely discovers such penalties, and its learned costs remain soft and opaque.   
Concretely, in ReCoDe, each agent’s \textit{learned} policy $\pi:\mathcal{O}\!\to\!\Theta$ maps its local observation $\mathbf{o}(t)$ to the parameters $\theta(t)=(\mathbf{a}(t),b(t))$ of a \emph{single quadratic constraint}
\(
\|\,\mathbf{u}(t)-\mathbf{a}(t)\|_2 \;\le\; b(t),
\)
added to the expert controller at every timestep.  
Here $\mathbf{u}(t)\!\in\!\mathbb{R}^m$ is the constrained optimization's decision variable, $\mathbf{a}(t)\!\in\!\mathbb{R}^m$ is a \textit{reference action} suggested by the policy, and the \textit{uncertainty radius} $b(t)\!\in\!\mathbb{R}_{\ge 0}$ dictates how tightly the solver must follow that reference—the larger $b(t)$, the more the agent defers to the original controller.  Thus the action space of the policy consists of the constraint parameters $(\mathbf{a}(t),b(t))$, rather than the actual control input $\mathbf{u}$, which is outputted by the solver.


\subheadline{Scope and Results.} 
We investigate ReCoDe’s mechanism theoretically and empirically. On the theory side we show that enlarging the uncertainty radius $b(t)$--thus increasing the agent's reliance on the default controller--can improve performance  in some situations (Proposition~\ref{thm:uncertaintyestimate}), while shrinking it enables precise learned control  (Propostion~\ref{thm:trajectory_tracking}).  
Empirically we confirm that ReCoDe learns to exploit this: it tightens its quadratic constraint in congested situations requiring precision and relaxes it when the path is clear, shifting authority between the learned policy and the default controller based on the situation (Figures~\ref{fig:b_vs_y_position}--\ref{fig:b_vs_num_neighbors}).  
This supports our claims about the benefits of retaining the default controller (and in particular, its optimization objective). 
We perform an ablation to further compare learning only constraints, learning only the objective, and learning both, and find that learning \emph{just} the constraint parameters yields the highest return and fastest convergence (Figure~\ref{fig:constraintobjectiveablation}).

While ReCoDe is a general method, we focus on studying its effectiveness in navigation and consensus settings. We design four experiments where coordination is necessary due to the small size of the environment, formation constraints, or conflicting incentives. We compare ReCoDe's performance in these experiments to several baselines: RVO \cite{van2008reciprocalvelocityobstaclesRVO}, a gold-standard non-learning methods for collision avoidance; two different hybrid methods from the literature \cite{gao2023onlinecbf,alshiekh2018safeshielding}; and end-to-end MARL. Across all four scenarios, ReCoDe attains, on average, 18\% better reward than the next-best method (Table \ref{fig:experiment scenarios and results}), and trains using just 5\% as many samples as end-to-end MARL (Figure \ref{fig:sample_efficiency}). We further demonstrate that ReCoDe preserves safety throughout training and deployment, and find that ReCoDe's learned constraints still provide a benefit to the handcrafted controller even in less coordination-heavy tasks (Figure \ref{fig:complexityvsreward}). Finally, we deploy ReCoDe on real robots, tasking two robot teams in a narrow corridor to swap positions. We find that the default controller often fails this task, leading robots into a deadlock, whereas ReCoDe's additional constraints let robots coordinate to avoid deadlock and successfully swap (Figure \ref{fig:robomaster_exp}). 




\textbf{Related Work.}  Classical multi-robot control is rooted in \emph{constrained optimization}, where an objective encodes the mission and constraints enforce safety, kinematics and environmental limits \citep{gregory2018constrained}. Linear programming and quadratic programming are the most common type of optimization \citep{earl2002modeling,fallahi2019linear}, often making use of control-barrier or Lyapunov functions for analytic collision avoidance and tracking guarantees  \citep{nocedal2006quadratic,nguyen2016exponential,gao2023onlinecbf}. Non-convex solvers can handle richer dynamics at higher computational cost \citep{tran2017nonlinear}. In contrast, \emph{multi-agent learning} approaches—including MARL—optimize neural policies directly from interaction data and have been applied to many  coordination problems \citep{busoniu2008comprehensive,chu2019multi,xue2023multi,gronauer2022multi}.  In MARL, agents learn to make decisions by interacting with the environment and updating their policies based on the feedback. Unlike constrained optimization, these methods do not require an analytic formalization of the desired task, and are more adaptible. However, they sacrifice the analytic guarantees of model-based controllers and may converge slowly when safe, high-reward actions are sparse. We provide a more detailed survey of both strands in Appendix \ref{appendix:extendedrelatedworks}.

\textit{Hybrid methods} combine MARL with constrained optimization. Relevant works in this space are \cite{gao2023onlinecbf}, \cite{romero2024actorcriticmodelpredictivecontrol}, and  \cite{alshiekh2018safeshielding}. In~\cite{gao2023onlinecbf}, MARL is used to optimize parameters of CBFs in constrained optimization problems for multi-agent navigation. In this work, rather than optimizing existing constraints, we learn additional, entirely new, constraints. The methods in \cite{alshiekh2018safeshielding} and  \cite{romero2024actorcriticmodelpredictivecontrol} can be seen as shaping the \textit{objective} of a constrained optimization problem, in different ways. While not applicable to our evaluated experimental scenarios, in \cite{romero2024actorcriticmodelpredictivecontrol}, RL is used to learn the objective function of a model-predictive control system for a single agent, under analytic assumptions that enable gradient backpropagation. In \cite{alshiekh2018safeshielding}, the \textit{shielding} method is introduced, which intervenes when the learned policy’s action violates safety constraints. One implementation of shielding uses constrained optimization whose \textit{objective} is to find a safe action closest to the policy's. ReCoDe instead attempts to generate an action based on a user-provided objective function and safety constraints, and generates additional constraints to further guide this optimization. We compare to shielding and Online CBF, and find that ReCoDe outperforms both these methods in a variety of navigation tasks (Section \ref{section:evaluation}). 








\vspace{-1mm}
\section{Problem Setting and Method: Dynamic Constraint Design}
\vspace{-1mm}

\textbf{Setting.} We consider a multi-agent system consisting of \(N\) agents in a shared environment. The primary objective of each agent is to maximize its cumulative reward over time. The reward at time $t$ reflects the agent's performance in the environment and may depend on factors such as task completion, efficiency, or cooperation with other agents. Agents interact with their environment through control inputs, which are obtained by solving an optimization problem whose constraints the agent itself can partially specify through choosing parameters $\theta(t)$. This optimization problem gives \textit{instantaneous control input} at every time step (i.e., there is no MPC-like receding horizon).

Let \(\mathbf{x}_i(t) \in \mathbb{R}^{n}\) denote the state of agent \(i\) at time \(t\), and \(\mathbf{u}_i(t) \in \mathbb{R}^{m}\) denote its control input. The dynamics of our agents are given by the discrete system $\mathbf{x}_i(t+dt) = f_i\left( \mathbf{x}_i(t), \mathbf{u}_i(t) \right)$, where \(f_i\) represents the dynamics of agent \(i\) and $dt$ is a constant representing the length of a time step.  Furthermore, at each time \(t\), agent \(i\) receives an observation \(\mathbf{o}_i(t) \in \mathcal{O}_i\), which depends on its own state, the states of other agents (if observed), and environmental variables: $\mathbf{o}_i(t) = h_i\left( \mathbf{x}_i(t), \mathbf{x}_{-i}(t), \mathbf{e}(t) \right)$,  where \(h_i\) is the observation function, \(\mathbf{x}_{-i}(t)\) denotes the states of other visible agents, and \(\mathbf{e}(t)\) represents external environmental factors. Given the observation and constraint parameters \(\boldsymbol{\theta}_i(t)\), agent \(i\) thus solves the constrained optimization problem
\(
\min_{\mathbf{u}_i(t)\in\mathcal{U}_i(\mathbf{o}_i(t);\boldsymbol{\theta}_i(t))}
J_i\!\left(\mathbf{o}_i(t),\mathbf{u}_i(t)\right)
\),
where
\(\mathcal{U}_i:=\{\mathbf{u}_i(t)\mid
g_k(\mathbf{o}_i(t),\mathbf{u}_i(t);\boldsymbol{\theta}_i(t))\le 0,\;k=1,\dots,K\}\).



Here, \(J_i\left(  \mathbf{o}_i(t), \mathbf{u}_i(t)\right)\) is a strictly convex in $u$, quadratic cost function, possibly representing factors like energy expenditure or deviation from a desired trajectory.
$\mathcal{U}_i$ is the set of admissible controls parametrized by $\theta$, where \(g_i\) represents the constraint functions. The constraint parameters of agent $i$ are selected based on its current observation \(\mathbf{o}_i(t)\) and its policy: $\boldsymbol{\theta}_i(t) = \pi_i\left( \mathbf{o}_i(t) \right)$, where \(\pi_i: \mathcal{O}_i \rightarrow \Theta_i\) is a policy mapping observations to parameters.

%

%

\textbf{Method.}
\label{section:method}
To enable agents to design constraints, we propose  (Re)inforcement-based (Co)nstraint (De)sign. ReCoDe trains agents in simulation using MAPPO, an actor-critic MARL algorithm \citep{ning2024survey}. Each agent $i$ seeks to learn a policy \(\pi_i\) that maps observations to constraint parameters, aiming to maximize the agent's reward. We design each agent's policy network to leverage relational information in the multi-agent system through a mechanism that aggregates nearby agents' messages. Although any such mechanism could work in principle, we elect to use Graph Neural Networks (GNNs). GNN architectures enable decentralized execution in inference time through message passing, where each agent computes messages \(m_{ij}\) to send to neighbors; aggregates incoming messages \(m_i = \sum_{j \in \mathcal{N}_i} m_{ij}\); and updates its state \(\mathbf{n}_i' = \sigma(m_i)\). This local computation ensures that each agent is decentralized, relying only on its own and neighbors' information. The unique structure of GNNs make the agent's perception of its neighborhood both \textit{permutation invariant} and \textit{dynamic}. That is, the order of neighboring agents does not affect the computation, and the neighborhood can adapt to external constraints, such as a limited sensing range. An overview of our method is shown in Figure \ref{fig:robomaster_exp}. Further implementation details are available in Appendix \ref{appendix:implementation-details}.

\headline{Centralized Training, Decentralized Execution}. We adopt a CTDE setup to speed up data collection during training \citep{amato2024introduction}. Specifically, we run \(M\) environment instances in parallel, each with \(N\) agents, and aggregate the resulting \(\mathcal{O}(M \times N)\) optimization problems into a \emph{single batched program}. However, if just one agent learns a parameter configuration that makes its constraints infeasible, the entire batch solver can fail, making it difficult to identify which agent caused the issue. To circumvent this, we introduce a \emph{slack variable} \(s_k\) only in \textit{learned} constraints (thus safety is unaffected), transforming $\mathcal{U}_i$ into $\mathcal{U}_i^{s} := \left\{ \mathbf{u}_i(t) \ \bigg| \ g_k\left( \mathbf{o}_i(t), \mathbf{u}_i(t); \boldsymbol{\theta}_i(t) \right) \leq s_k, k \in [1,K] \right\}$. During training, each agent  tries to minimize $J_i\left( \mathbf{o}_i(t), \mathbf{u}_i(t) \right) + \sum_{k=1}^{|\mathcal{U}_i^{s}|} \mathbf{\lambda}_k s_k$
where each slack \(s_k \geq 0 \) is heavily penalized (\(\lambda_k \gg 0\)), so it is only nonzero when no feasible solution exists. This setup also identifies infeasible programs by flagging the agents whose \(s_k\) is nonzero. This centralization takes place in \emph{training}; at deployment, each agent is fully decentralized, and solves its own local optimization problem with the final learned constraints based only on local observations and communication.

\headline{Constraint Form.} In ReCoDe, agents augment a default controller with additional constraints for improved performance. We focus on learning a single, quadratic constraint $\| \mathbf{u}_i(t) - \mathbf{a}_i(t) \|_2 \leq b_i(t) + s_0$ parametrized by $\boldsymbol{\theta}_i(t)=(\mathbf{a}_i(t),b_i(t))$, where $\mathbf{a}_i(t) \in \mathbb{R}^m$ is a \textit{reference action} suggested by the policy, $b_i(t) \in \mathbb{R}_{\geq 0}$ is a radius that decides how much the handcrafted controller can steer away from $\mathbf{a}(t)$, and $s_0$ is a slack variable. We explain the reasoning for this, and consider other types of constraints, in Appendices   \ref{appendix:whynotlinearconstraints} and \ref{appendix:differentconstraintforms}. The learned agent policy $\pi_i:\mathcal{O}_i\!\to\!\Theta_i$ maps its observation $\mathbf{o}_i(t)$ to the constraint parameters $\boldsymbol{\theta}_i(t)=(\mathbf{a}_i(t),b_i(t))$, and agent $i$ solves:

\vspace{-4mm}
\begin{equation}
\begin{aligned}
\min_{\mathbf{u}_i(t)} \ & J_i\left( \mathbf{o}_i(t), \mathbf{u}_i(t)  \right) + \lambda_0 s_0 
\text{\ \ s.t.} & \left\| \mathbf{u}_i(t) - \mathbf{a}_i(t) \right\|_2 \leq b_i(t) + s_0, \mathbf{u}_i(t) \in \mathcal{U}_i^{s}\left( \mathbf{o}_i(t)\right)\text{.}
\end{aligned}
\label{eq:final_optimization_problem}
\end{equation}
\vspace{-4mm}

Here, $\mathcal{U}_i^{s}\left( \mathbf{o}_i(t)\right)$ defines the constraints of our default (expert) controller, which may be parametrized by agent observations but, unlike $(\mathbf{a}(t),b(t))$, not \textit{learned} by our algorithm. We call $b_i(t)$ the \textit{uncertainty radius} of agent $i$ at time $t$, since it controls how strict the learned constraint is. A larger value of $b_i(t)$ can be viewed as the learned policy having uncertainty about how optimal the action $\mathbf{a_i}(t)$ is, hence preferring to influence the default controller less.


We \textit{assume} that in \eqref{eq:final_optimization_problem}
the objective $J_i$ is \emph{strictly convex} in $\mathbf u$
and that, for every observation $\mathbf o$ and parameter vector
$\boldsymbol\theta\!\in\!\Theta$, the feasible set is \emph{non‑empty and convex}.
We further assume that the problem \eqref{eq:final_optimization_problem}
admits a \emph{unique} minimizer
$\mathbf u^{\star}(\mathbf o,\boldsymbol\theta)$,
and that the solution mapping
$\boldsymbol\theta\mapsto\mathbf u^{\star}(\mathbf o,\boldsymbol\theta)$
is continuously differentiable in a neighborhood of each
$\boldsymbol\theta$
(\cite{fiacco1983introduction} lists mild regularity
conditions under which this property holds). Because $J_i(\mathbf o,\mathbf u)$ is strictly convex
quadratic in~$\mathbf u$ and constraints are convex, 
\eqref{eq:final_optimization_problem} is a \emph{convex QCQP}. Such problems are known to be efficiently solvable \cite{lobo1998applications}. This is important for inference and training, as we must deploy many instances of \eqref{eq:final_optimization_problem} during data collection.

\section{Analysis}
\label{section:analysis}

Our analysis shows that tightening the uncertainty radius $b(t)$ enables \textit{adaptability}, letting ReCoDe track any safe trajectory, whereas enlarging the radius lets the handcrafted controller take over and raise the reward when the policy is uncertain—thereby validating ReCoDe’s design of dynamically balancing learned and expert control (proofs in Appendices ~\ref{appendix:proofoftrajectorysafety}--\ref{appendix:proofofuncertaintyestimate}).

\headline{Adaptability.}
Since ReCoDe uses a constrained-optimization framework, user-defined safety constraints are never violated. We show here that as long as it remains within these constraints, and the slack penalty~\(\lambda_0\) is sufficiently large, ReCoDe is  \textit{precise and adaptable}: the agent can choose constraints that force its controller to track any safe, feasible trajectory with arbitrarily small error. Formally, let $T \in \mathbb{N}$ be a finite time horizon. Consider an agent in our system whose initial state is $x^*(1)$, and a feasible desired trajectory of actions and states  $(x^*(1), u^*(1)), (x^*(2), u^*(2)), \dots, (x^*(T), u^*(T))$ where executing $u^*(t)$ in state $x^*(t)$ takes the agent to state $x^*(t+1)$. Let $B_\varepsilon(p) = \{y \in \mathbb{R}^m : \|y - p\| < \epsilon\}$ be the $\varepsilon$-neighbourhood of a point $p$. Assume that, for each time \(t\), the desired action \(u^*(t)\) is strictly feasible under the handcrafted constraints without requiring any slack, i.e., with \(s_0(t)=0\). More formally, there exists \(\eta > 0\) such that for all \(t\), $B_{\eta}(u^*(t)) \subseteq \mathcal{U}_i^{s}(\mathbf{o}_i(t))$, where \(\mathcal{U}_i^{s}(\cdot)\) denotes the feasible set defined by the hand-crafted constraints with slack parameters, and at \(u^*(t)\) we have \(s_0(t)=0\). Then we have:

\begin{proposition}[$\varepsilon$-Close Trajectory Tracking]
\label{thm:trajectory_tracking}
For any \(\varepsilon > 0\), there exists a  sufficiently large penalty factor \(\lambda_0\) and a sequence of learned constraint parameters \(\{(\mathbf{a}(t), b(t))\}_{t=1}^T\), with \(\mathbf{a}(t) \in \mathbb{R}^m\) and \(b(t)\in \mathbb{R}_{> 0}\), such that the unique optimal solutions \(u^{\text{opt}}(t)\) of the optimization problem \eqref{eq:final_optimization_problem} satisfy
 $\|u^{\text{opt}}(t) - u^*(t)\| \leq \varepsilon, \quad \forall t=1,\dots,T$.  Moreover, the resulting state trajectory \(\{x(t)\}_{t=1}^T\) satisfies: $
\|x(t) - x^*(t)\| \leq \varepsilon, \quad \forall t=1,\dots,T$.
\end{proposition}

\headline{Uncertainty Mitigation.} In many RL algorithms, such as actor-critic, an agent wants to pick
a control input \(u\) to maximize the \emph{true} expected reward 
\(Q_i^{*}(\mathbf o,u)\) given observation $\mathbf o$, but must optimize some imperfect learned proxy of $Q_i^*$--a critic--which we denote \(Q_i^{l}(\mathbf o,u)\).
ReCoDe, instead, has the agent policy output
 \(\mathbf a(\mathbf o)\) and an \emph{uncertainty radius} \(b(\mathbf o)\)
that define a ball (constraint) where $u$ should lie   
\(\|u-\mathbf a(\mathbf o)\|_{2}\le b(\mathbf o)\).
We make the somewhat simplifying \textit{interpretation} that the agent chooses \(\mathbf a(\mathbf o)\) because it maximizes \(Q_i^{l}(\mathbf o,u)\) for some learned proxy $Q_i^{l}$. Next, the handcrafted optimization objective
\(J_i\bigl(\mathbf o,u\bigr)\) picks $u$ 
\emph{inside} the ball; the larger the radius $b(\mathbf o)$, the stronger the expert
controller’s influence. We will show it is beneficial to enlarge $b(\mathbf o)$, i.e., to mix the expert controller with the learned policy, when \(Q_i^{l}(\mathbf o,u)\) is more \textit{uncertain} than $J_i$ given observation $\mathbf o$. By uncertain we mean that \(Q_i^{l}(\mathbf o,u)\)'s gradient is locally bounded by a small \(\delta\) and therefore it assigns roughly the same value to all actions locally\footnote{Strictly speaking, flatness does not necessarily imply uncertainty; we use it as an imperfect proxy.}. The result requires that for some $c_1(\mathbf o), c_2(\mathbf o) > 0$, in a neighborhood of \(\mathbf a(\mathbf o)\), a weighted
combination \(c_1 Q_i^{l}-c_2 J_i\) is a good approximation of 
 \(Q_i^{*}\).\footnote{If \(c_1 Q_i^{l}-c_2 J_i\) locally approximates \(Q_i^{*}+c_3\) for some constant $c_3(\mathbf o)$, a similar result holds. Note: this approximation assumption is weaker than assuming $Q_i^l$ or $J_i$ on their own can be used to approximate $Q_i^*$.}

\begin{proposition}
\label{thm:uncertaintyestimate}
Assume there is $r>0$ such that for all actions
$u\in B_{r}\!\bigl(a(\mathbf{o})\bigr)$ we have
$\bigl|Q^*(\mathbf{o},u)
      -\bigl[c_1 Q_i^l(\mathbf{o},u)-c_2 J_i(\mathbf{o},u)\bigr]\bigr|
      \le \varepsilon$
and
$\|\nabla_u Q_i^l(\mathbf{o},u)\|_2\le\delta_1$.
Assume also that there exists a \emph{unit} direction $\mathbf d$ and a
constant $\delta_2>\delta_1$ such that
\(
  \mathbf d^{\!\top}\nabla_u\!\bigl[-J_i\bigr]
  \bigl(\mathbf{o},\mathbf{a}(\mathbf{o})+x\mathbf d\bigr)
  \;\ge\;\delta_2
  \quad\forall\,x\in[0,r].
\)
Let $c_2\delta_2-c_1\delta_1=\Delta$.
If every action $u\in B_{r}\!\bigl(a(\mathbf{o})\bigr)$ is strictly in
problem~\eqref{eq:final_optimization_problem}'s feasible action set  without slack and
$\lambda_0$ (the slack penalty on $s_0$) is sufficiently large, then
\(
   Q^*(\mathbf{o},u^{\mathrm{opt}}(\mathbf{o}))
   \;\ge\;
   Q^*(\mathbf{o},a(\mathbf{o})) + r\Delta - 2\varepsilon,
\)
where $u^{\mathrm{opt}}$ is the solution to
\eqref{eq:final_optimization_problem} (i.e., ReCoDe's output) given $b(\mathbf o) = r$.
\end{proposition}


\subheadline{How to read the bound.} \(r=b(\mathbf o)\) is the \emph{uncertainty radius}; enlarging it lets the
handcrafted objective \(J_i\) shape the solver’s choice inside a wider ball.
The constant \(\delta_1\) upper-bounds the gradient of 
\(Q_i^{l}\) in that ball, so it quantifies how \emph{flat}—hence how
uncertain—the learned policy is locally.
By contrast \(\delta_2 > \delta_1\) lower-bounds the directional derivative of
\(-J_i\) along \emph{at least one} direction, certifying that the expert
objective is less flat than $Q_i^l$.
If 
\(c_2\delta_2>c_1\delta_1\), then \(\Delta=c_2\delta_2-c_1\delta_1>0\);
consequently the term \(r\Delta\) in the bound is positive and the solver
finds an action that beats the critic’s own maximizer by at least
\(r\Delta\!-\!2\varepsilon\).
In other words, when, locally, the critic is flat but the expert objective is
decisive, enlarging the uncertainty radius \emph{mitigates uncertainty} by mixing expert
and learned knowledge.

The proposition does not directly imply a strategy, as \(c_{1}(\mathbf o)\) and \(c_{2}(\mathbf o)\) are \emph{unknown}.  Instead, it explains \emph{why} giving the policy control over the uncertainty radius can be useful. To properly determine $b(\mathbf o)$, the an agent should have some notion of how to weigh the handcrafted vs. learned optimization goal locally. Does ReCoDe  tune $b(\mathbf o)$ like this in practice? In Section~\ref{section:evaluation} we give empirical evidence for this: \(b(\mathbf o)\) \emph{shrinks} in crowded, high-interaction states where the learned policy is more reliable than the expert, but  \emph{expands} once the path is clear, leaning on the handcrafted
controller.




\section{Evaluation}
\label{section:evaluation}
\vspace{-1mm}

We evaluate ReCoDe in four \textit{multi-agent navigation and consensus} tasks that are designed to expose two common failure modes in multi-robot control. The first mode appears when safe, reward-producing actions are sparse; in such settings pure reinforcement learning spends most of its time exploring moves that end in collisions and learns very slowly. The second mode remains even when individual safe moves are plentiful and easy to compute: with several robots in close proximity, reciprocal blocking and group-level constraints can trap a system in deadlock, something a handcrafted controller often cannot anticipate or avert. 

The \textbf{Narrow Corridor} task places two teams at opposite ends of a narrow hallway and asks them to swap positions, so agents must discover when to yield in a space where successful moves are sparse. \textbf{Connectivity} requires a single team to navigate to the end of the hallway, but introduces static obstacles and and requires that every pair of robots stay within a fixed communication range, thereby preserving full connectivity throughout the task. This binds the motions of the entire group and requires them to collectively negotiate movements. The \textbf{Waypoint Navigation} scenario moves over-sized robots in a small room with random goals, frequently requiring robots to go ``the long way round'' to their location if deadlocks are to be avoided. Finally, the \textbf{Sensor Coverage} scenario is a multi-objective scenario that couples motion with high-level consensus about \textit{where} to go: a fleet of sensors, each assigned to monitor different phenomena, must collectively decide where to position themselves to attain the best overall coverage of the environment while never breaking their communication graph (see Appendix \ref{appendix:experiment_explanations}, Figure \ref{fig:sensorcoverage}). In all scenarios, holonomic agents observe their own position, distance to goal position, and relative position to obstacles and other agents within their communication range. Detailed scenario definitions are available in Appendix \ref{appendix:experiment_explanations}. 

Across all tasks we benchmark ReCoDe against multiple baselines. The first is the \textbf{handcrafted controller}—the best constraint-based controller we could craft for each scenario, the details of which we give in  Appendix \ref{appendix:experiment_explanations}. This is also the controller we use as a basis for ReCoDe. The other non-learning based method is \textbf{Reciprocal Velocity Obstacle} (RVO) algorithm \cite{van2008reciprocalvelocityobstaclesRVO}, a gold-standard method for multi-agent collision avoidance. To isolate the benefit of optimization we also include \textbf{Pure MARL}, an end-to-end MARL policy that directly controls the agents. Finally, we test against two hybrid methods, \textbf{Online-CBF} \citep{gao2023onlinecbf} and \textbf{shielding} \cite{alshiekh2018safeshielding}, covered in our Related Work section. In our implementation of shielding, the policy outputs a target velocity and passes it through a safety filter using the same safety constraints we used for ReCoDe; in Online CBF, we learn the parameter $k$ in the control barrier function of \eqref{eq:narrowcorridorcontroller}. In all experiments that make use of reinforcement learning, we used MAPPO and the same GNN-based actor-critic architecture introduced in Section \ref{section:method} (such architectures are SOTA in end-to-end MARL--see, e.g.,  \cite{nayak2023scalable}). We train all baselines for 7.2 million environment steps, or in the case of \textit{Pure MARL}, for longer until reward stabilizes.

\vspace{-2mm}
\subsection{Results}
\label{sec:result}
\vspace{-2mm}

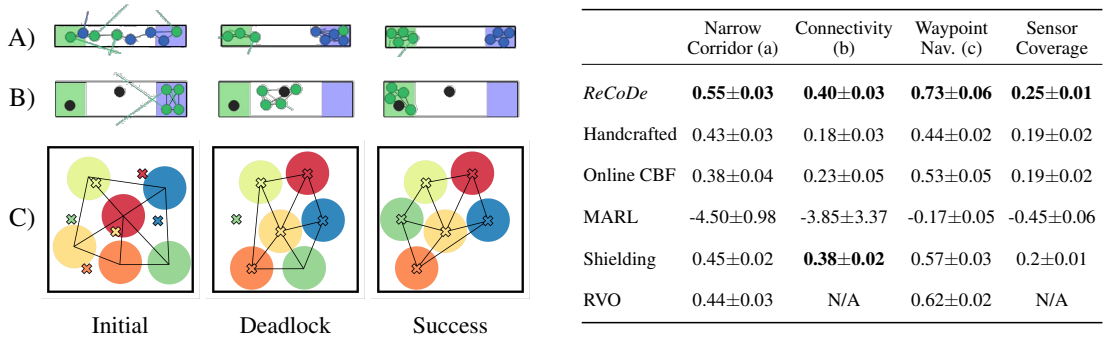
\begin{figure*}[t]
    \hspace*{-0.7cm}%
    \begin{minipage}{0.5\textwidth}
        \centering
        \begin{tikzpicture}
            \def\colwidth{0.3\textwidth}
            \def\verticaloffset{-8mm}
            \def\hspace{0.1cm}
            
            \coordinate (start) at (0,0);
            
            \node[anchor=east] at ($(start) +
            (-1.0,0)$) {A)};
            \node (a1) at ($(start) + (0,0.08)$) {\includegraphics[width=\colwidth]{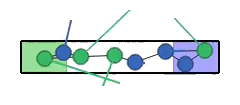}};
            \node (a2) at ($(start) + (\colwidth+\hspace,0)$) {\includegraphics[width=\colwidth]{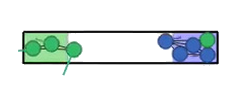}};
            \node (a3) at ($(start) + (2*\colwidth+2*\hspace,0.02)$) {\includegraphics[width=\colwidth]{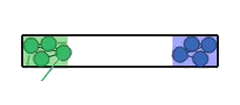}};
            
            \node[anchor=east] at ($(start) + (-1.0,\verticaloffset)$) {B)};
            \node (b1) at ($(start) + (0,\verticaloffset)$) {\includegraphics[width=\colwidth]{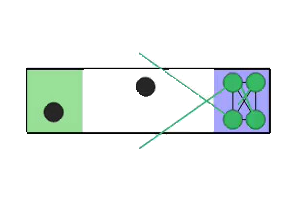}};
            \node (b2) at ($(start) + (\colwidth+\hspace,\verticaloffset)$) {\includegraphics[width=\colwidth]{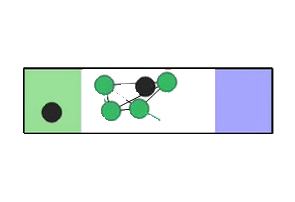}};
            \node (b3) at ($(start) + (2*\colwidth+2*\hspace,\verticaloffset)$) {\includegraphics[width=\colwidth]{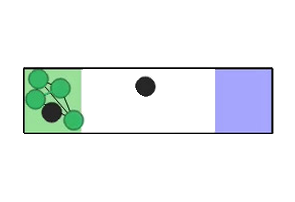}};
            
            \node[anchor=east] at ($(start) + (-1.0,3*\verticaloffset)$) {C)};
            \node (c1) at ($(start) + (0,3*\verticaloffset)$) {\includegraphics[width=\colwidth]{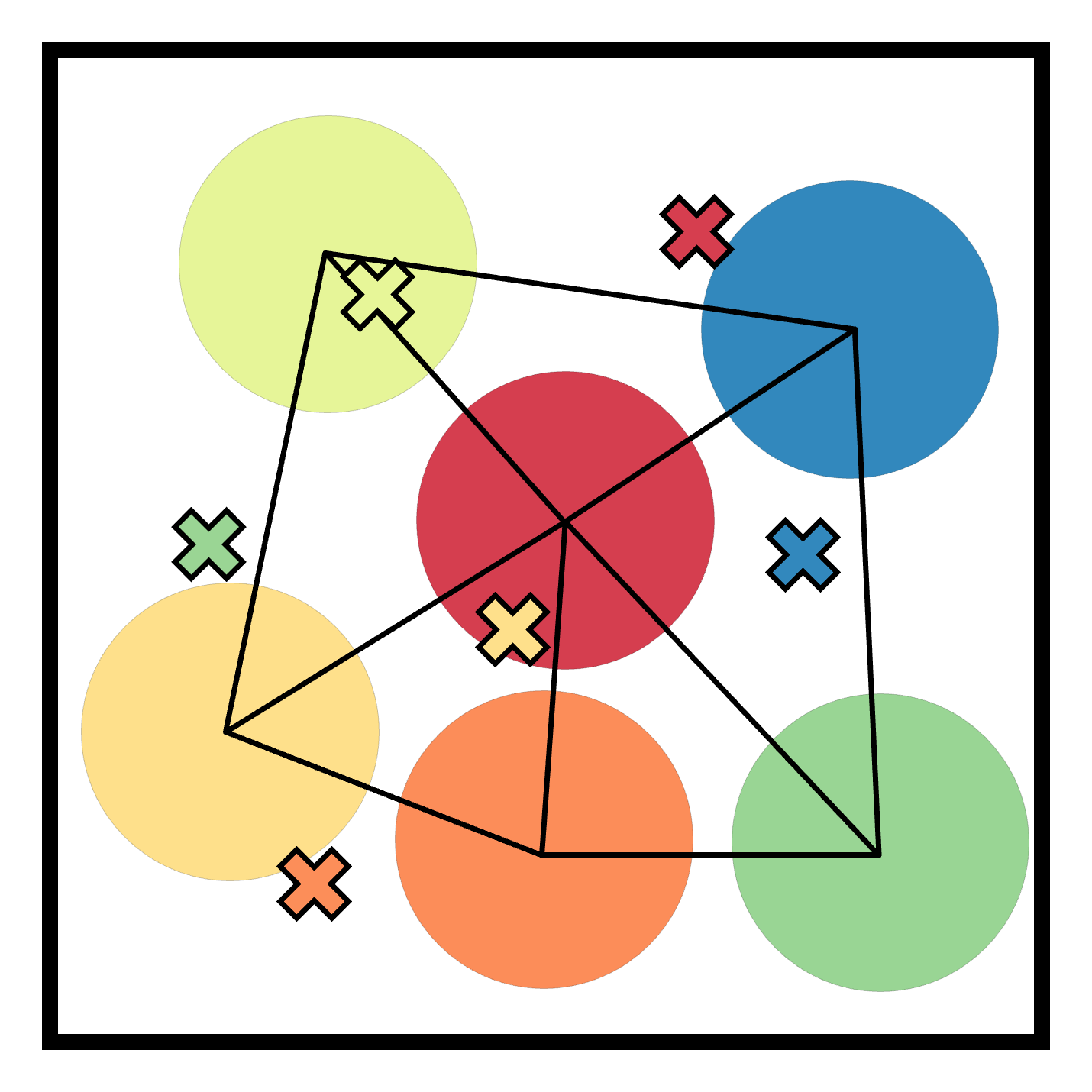}};
            \node (c2) at ($(start) + (\colwidth+\hspace,3*\verticaloffset)$) {\includegraphics[width=\colwidth]{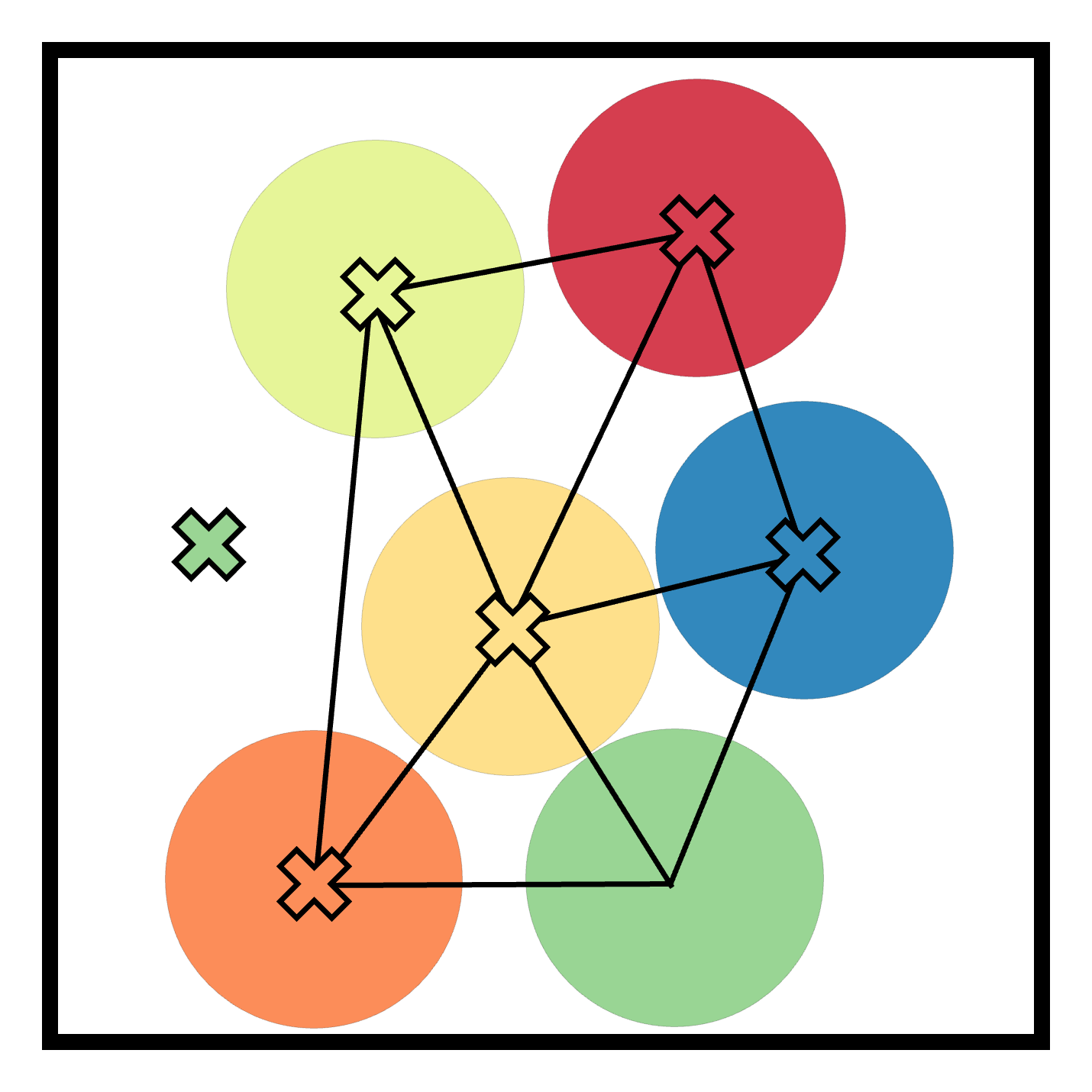}};
            \node (c3) at ($(start) + (2*\colwidth+2*\hspace,3*\verticaloffset)$) {\includegraphics[width=\colwidth]{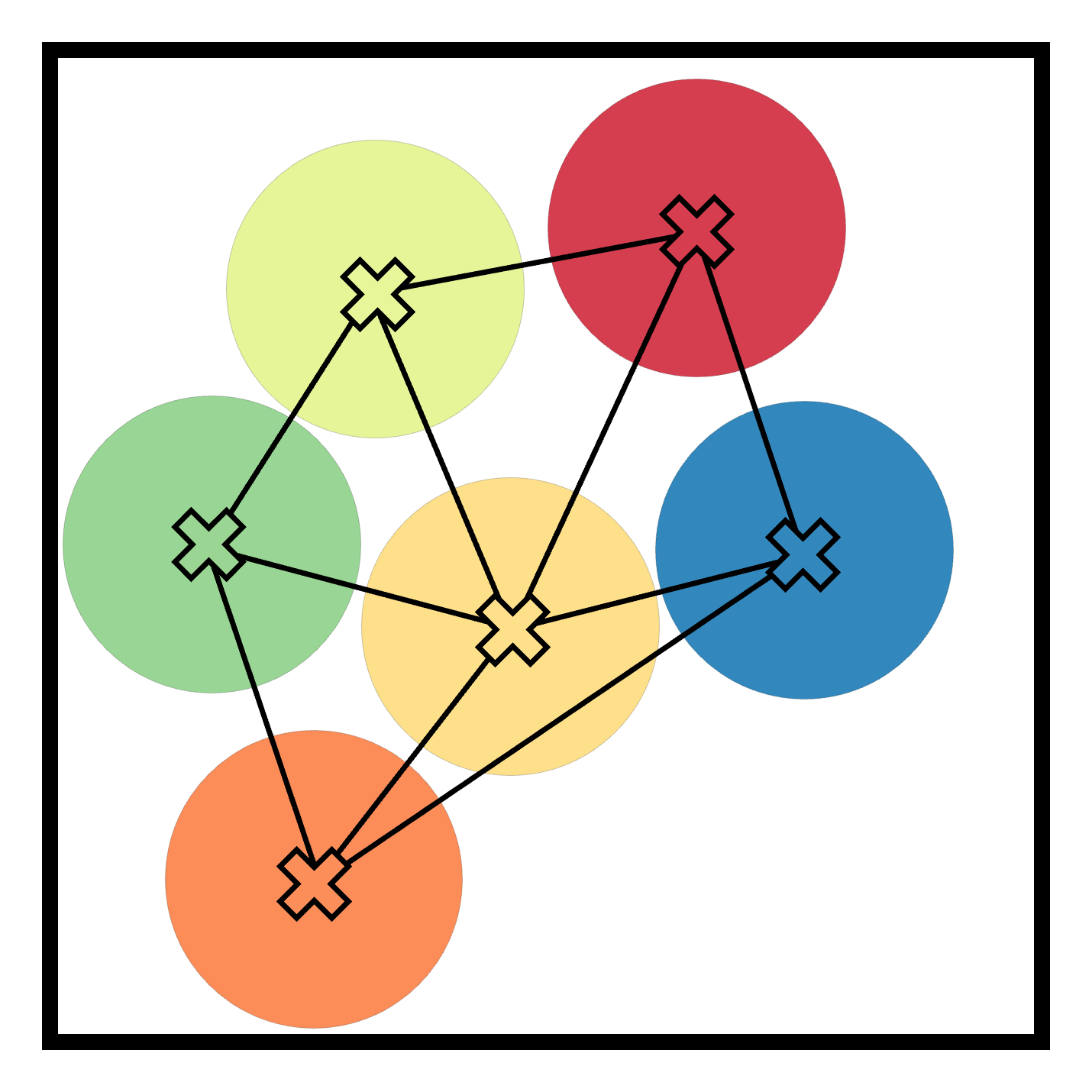}};
            
            \node[anchor=north] at ($(c1.south) + (0,-0.0)$) {\small Initial};
            \node[anchor=north] at ($(c2.south) + (0,-0.0)$) {\small Deadlock};
            \node[anchor=north] at ($(c3.south) + (0,-0.0)$) {\small Success};
        \end{tikzpicture}
    \end{minipage}%
    \hfill
    \begin{minipage}{0.5\textwidth}
        \setlength{\tabcolsep}{1pt}
        
        \renewcommand{\arraystretch}{2.0}
        \newcommand{\entry}[3]{
        \renewcommand{\arraystretch}{1.0}
        {#2$\pm$#3}
        }
        \newcommand{\bestentry}[3]{
        \renewcommand{\arraystretch}{1.0}
        {\textbf{#2$\pm$#3}}
        }    
        \newcommand{\std}[1]{$\pm#1$}
        
        \newcommand{\compactheader}[2]{
        \renewcommand{\arraystretch}{1.0}
        \begin{tabular}{c}#1\\#2
        \end{tabular}
        }
        
        \newcommand{\compactrowheader}[2]{%
            \setlength{\tabcolsep}{0pt}%
            \renewcommand{\arraystretch}{1.0}%
            \begin{tabular}[c]{@{}l@{}}#1\\#2%
            \end{tabular}%
        }
        
        \resizebox{\columnwidth}{!}{
        \begin{tabular}{lcccc}
        \toprule
         & \compactheader{Narrow}{Corridor (a)} & \compactheader{Connectivity}{(b)} & 
           \compactheader{Waypoint}{Nav. (c)} & 
           \compactheader{Sensor}{Coverage} \\ 
        \midrule
        \textit{{ReCoDe}} 
        & \bestentry{\textbf{0.59}}{0.55}{0.03}
        & \bestentry{\textbf{0.43}}{0.40}{0.03}
        & \bestentry{\textbf{0.81}}{0.73}{0.06}
        & \bestentry{\textbf{0.27}}{0.25}{0.01}
        \\
        \compactrowheader{Handcrafted}{}
        & \entry{0.48}{0.43}{0.03}
        & \entry{0.26}{0.18}{0.03}
        & \entry{0.47}{0.44}{0.02}
        & \entry{0.22}{0.19}{0.02}
        \\
        \compactrowheader{Online CBF}{}
        & \entry{0.45}{0.38}{0.04}
        & \entry{0.34}{0.23}{0.05}
        & \entry{0.63}{0.53}{0.05}
        & \entry{0.23}{0.19}{0.02}
        \\
        MARL
        & \entry{-3.60}{-4.50}{0.98}
        & \entry{-1.52}{-3.85}{3.37}
        & \entry{-0.09}{-0.17}{0.05}
        & \entry{-0.35}{-0.45}{0.06}
        \\
        Shielding
        & \entry{0.50}{0.45}{0.02}
        & \bestentry{0.42}{0.38}{0.02}
        & \entry{0.63}{0.57}{0.03}
        & \entry{0.22}{0.2}{0.01}
        \\
        RVO
        & \entry{0.48}{0.44}{0.03}
        & N/A
        & \entry{0.65}{0.62}{0.02}
        & N/A
        \\ 
        \bottomrule
        \end{tabular}%
        }
    \end{minipage}

    \caption{Experimental scenarios and results. Leftmost  column: initial conditions; middle: a possible deadlock scenario where agents require coordination to proceed; rightmost: scenario success. Agents are colored circles; black edges denote communication lines. (A) Narrow Corridor: blue/green agents navigate to color-matched regions while communicating with neighbors. (B) Connectivity: agents bypass obstacles without breaking any communication links. (C) Waypoint Navigation: agents coordinate to reach color-matched goals, requiring some to leave goals despite short-term incentives temporarily. The table shows average rewards per time step for ReCoDe, handcrafted constraints, Online CBF, RVO, shielding, and Pure MARL across 75 random starting conditions (mean $\pm$ standard deviation over best 6 consecutive training steps). Maximum possible rewards are \textit{roughly} $1$ in Narrow Corridor, Connectivity, $1.5$ in Waypoint, and unknown in Sensor Coverage.}
    \label{fig:experiment scenarios and results}
    \vspace{-2mm}
\end{figure*}

 Our results are summarized in Table \ref{fig:experiment scenarios and results}. ReCoDe significantly outperforms the baselines in all tested scenarios, attaining, on average, 18\% greater reward than the next-best method. \textit{Pure MARL} performed the worst, and was unable to control agents and reach to an adequate performance across all scenarios. This is because our scenarios are very sensitive to small changes in the control input (e.g., in a narrow corridor a very small change in input makes the difference between a collision and successful navigation); it is difficult to obtain a positive reward without optimization. Qualitatively, the most frequent failure modes for other methods were deadlocks stemming from lack of coordination. 

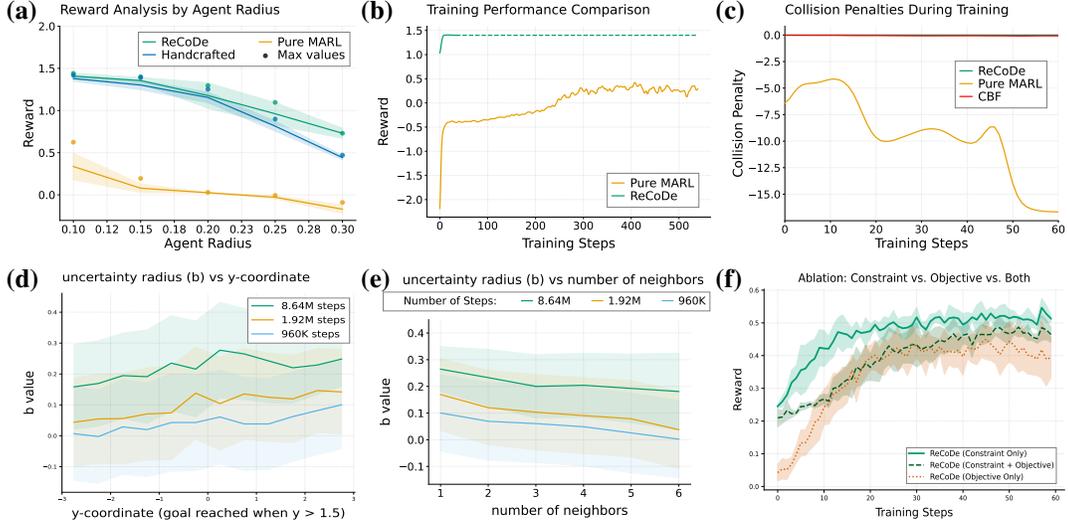
\begin{figure*}[t]
\centering
\begin{subfigure}[b]{0.325\textwidth}
  \centering
  \begin{tikzpicture}
    \node[anchor=south west,inner sep=0] (image) at (0,0){\includegraphics[width=\linewidth]{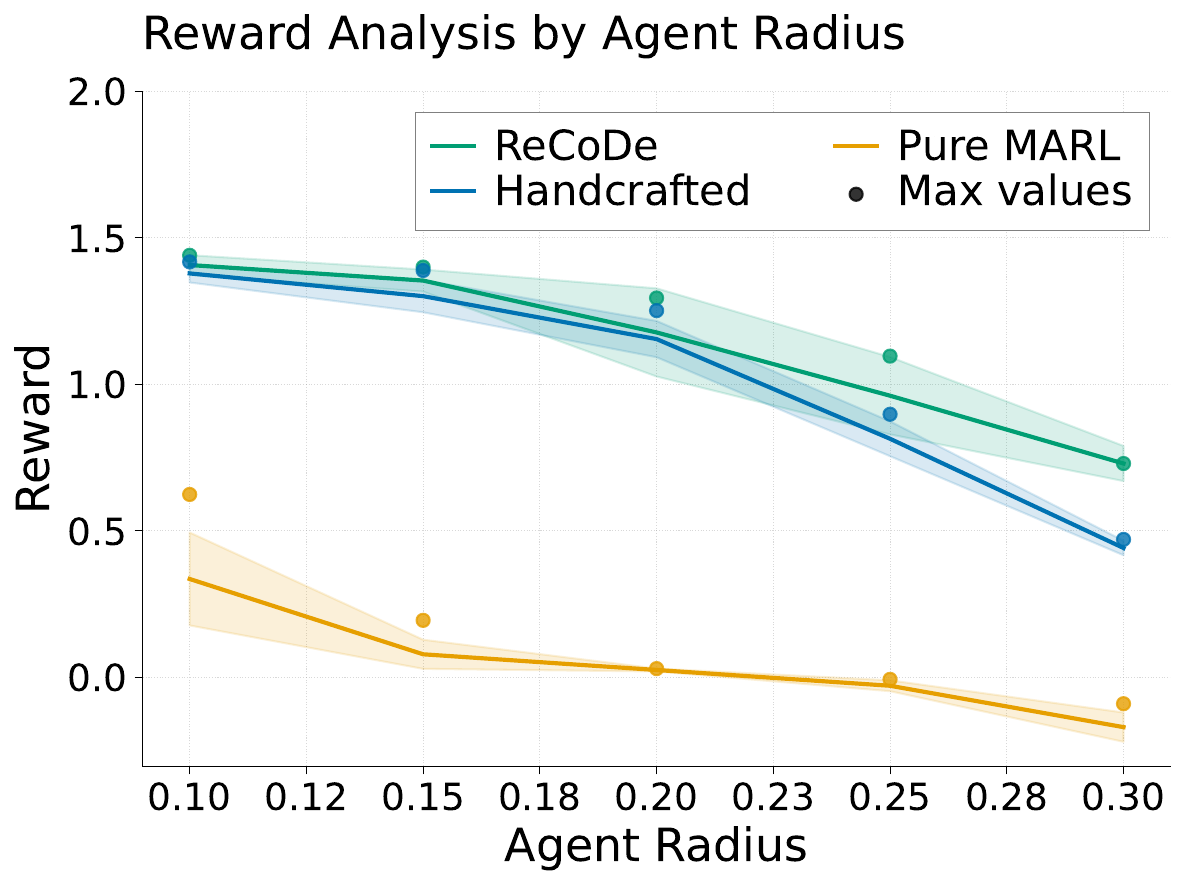}};
    \node[anchor=north west,xshift=-8pt,yshift=3pt] at (image.north west) {\textbf{(a)}};
  \end{tikzpicture}
  \phantomsubcaption
  \label{fig:complexityvsreward}
\end{subfigure}
\hfill
\begin{subfigure}[b]{0.325\textwidth}
  \centering
  \begin{tikzpicture}
    \node[anchor=south west,inner sep=0] (image) at (0,0)
      {\includegraphics[width=\linewidth]{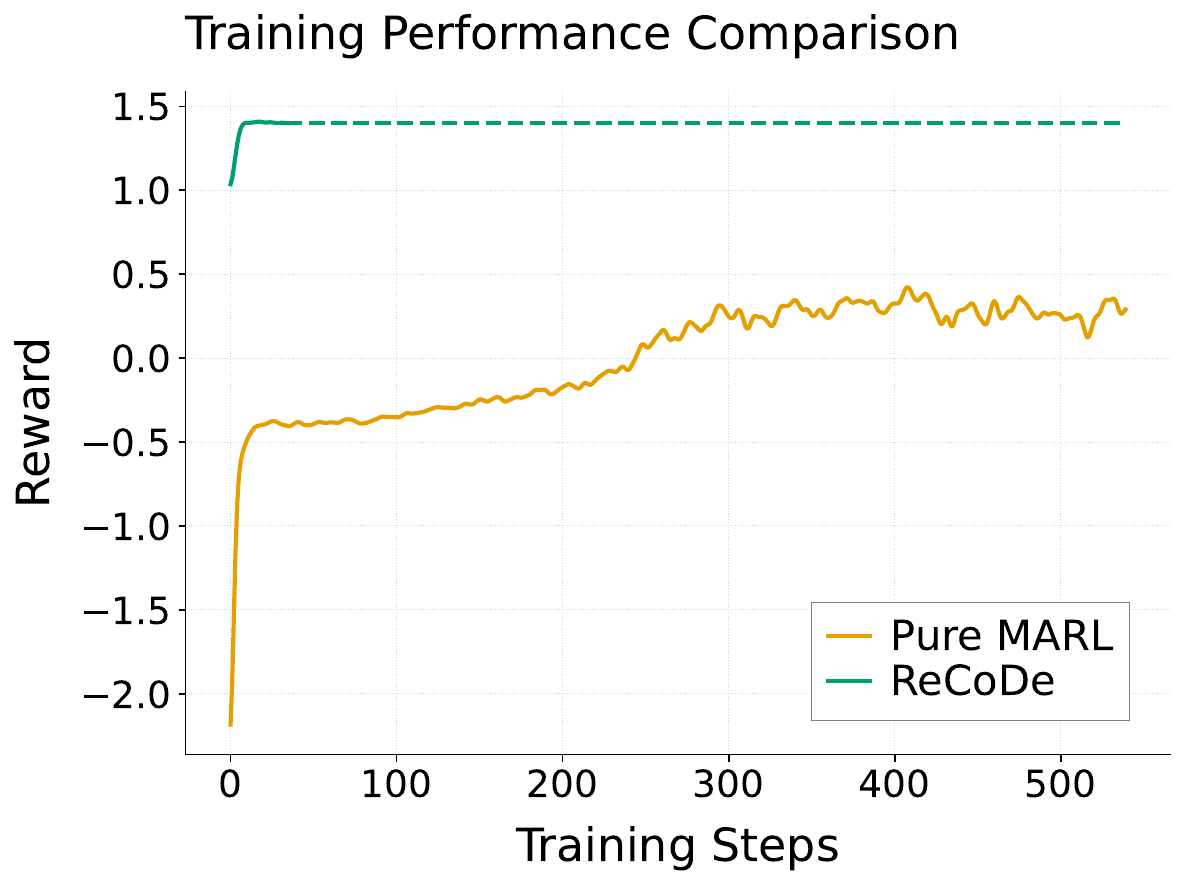}};
    \node[anchor=north west,xshift=-8pt,yshift=3pt] at (image.north west) {\textbf{(b)}};
  \end{tikzpicture}
  \phantomsubcaption
  \label{fig:sample_efficiency}
\end{subfigure}
\hfill
\begin{subfigure}[b]{0.325\textwidth}
  \centering
  \begin{tikzpicture}
    \node[anchor=south west,inner sep=0] (image) at (0,0)
      {\includegraphics[width=\linewidth]{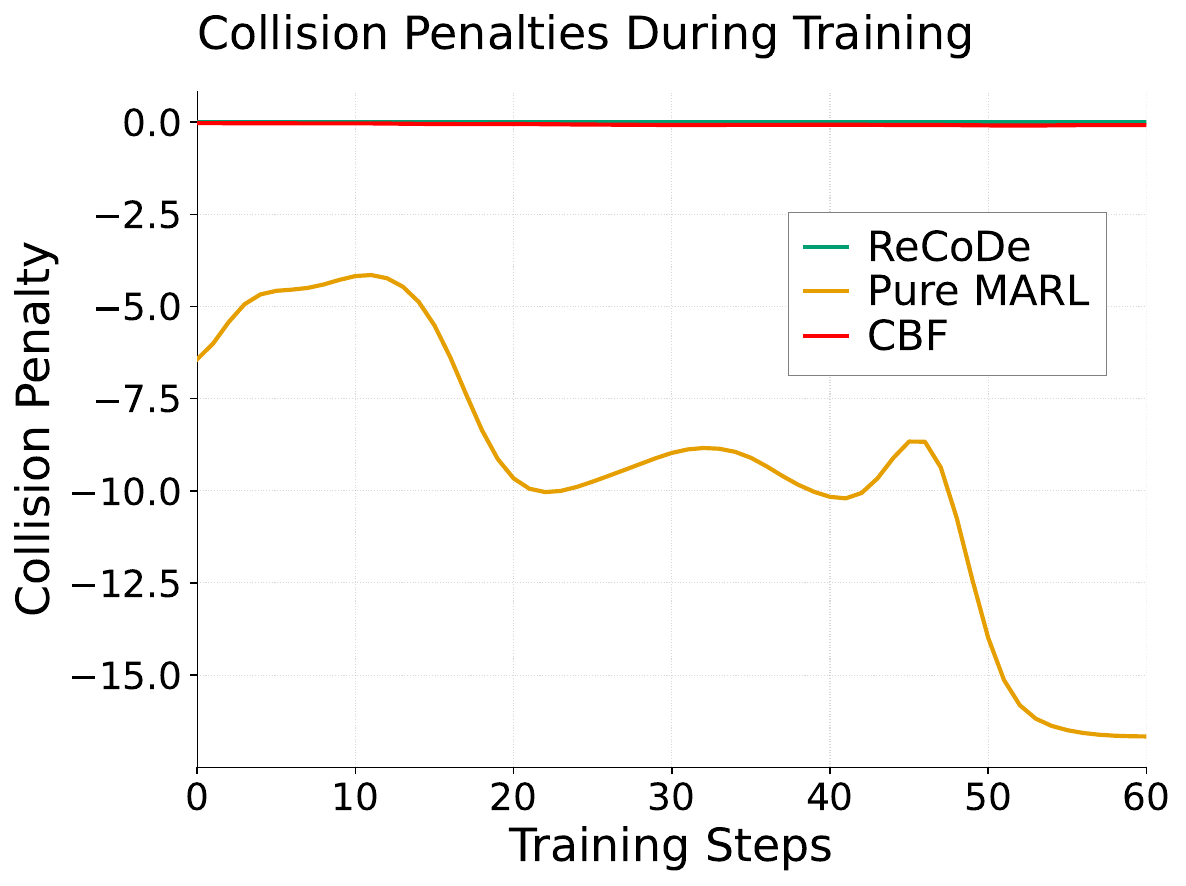}};
    \node[anchor=north west,xshift=-8pt,yshift=3pt] at (image.north west) {\textbf{(c)}};
  \end{tikzpicture}
  \phantomsubcaption
  \label{fig:collision_rew}
\end{subfigure}

\vspace{-1em} 

\begin{subfigure}[b]{0.325\textwidth}
  \centering
  \begin{tikzpicture}
    \node[anchor=south west,inner sep=0] (image) at (0,0)
      {\includegraphics[width=\linewidth,height=0.75\linewidth]{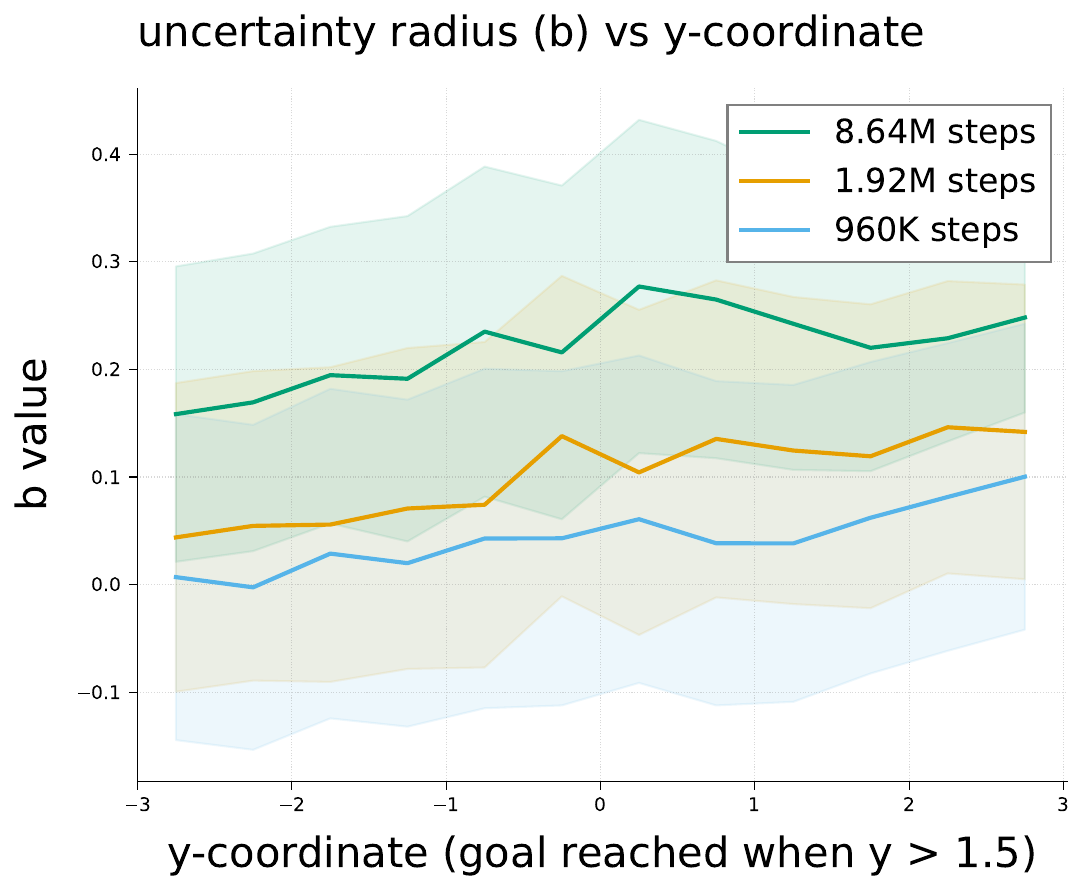}};
    \node[anchor=north west,xshift=-8pt,yshift=3pt] at (image.north west) {\textbf{(d)}};
  \end{tikzpicture}
  \phantomsubcaption
  \label{fig:b_vs_y_position}
\end{subfigure}
\hfill
\begin{subfigure}[b]{0.325\textwidth}
  \centering
  \begin{tikzpicture}
    \node[anchor=south west,inner sep=0] (image) at (0,0)
      {\includegraphics[width=\linewidth,height=0.75\linewidth]{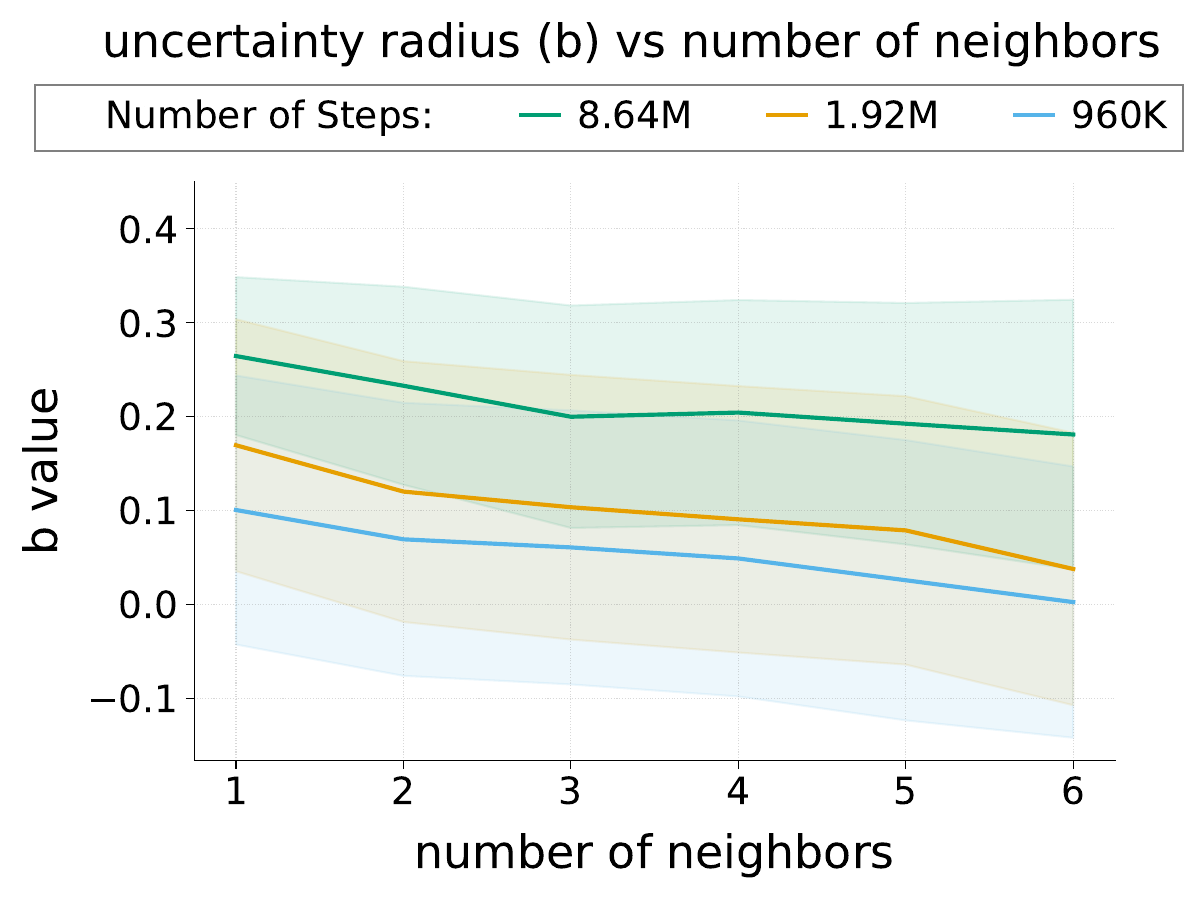}};
    \node[anchor=north west,xshift=-8pt,yshift=3pt] at (image.north west) {\textbf{(e)}};
  \end{tikzpicture}
  \phantomsubcaption
  \label{fig:b_vs_num_neighbors}
\end{subfigure}
\hfill
\begin{subfigure}[b]{0.325\textwidth}
  \centering
  \begin{tikzpicture}
    \node[anchor=south west,inner sep=0] (image) at (0,0)
      {\includegraphics[width=\linewidth,height=0.75\linewidth]{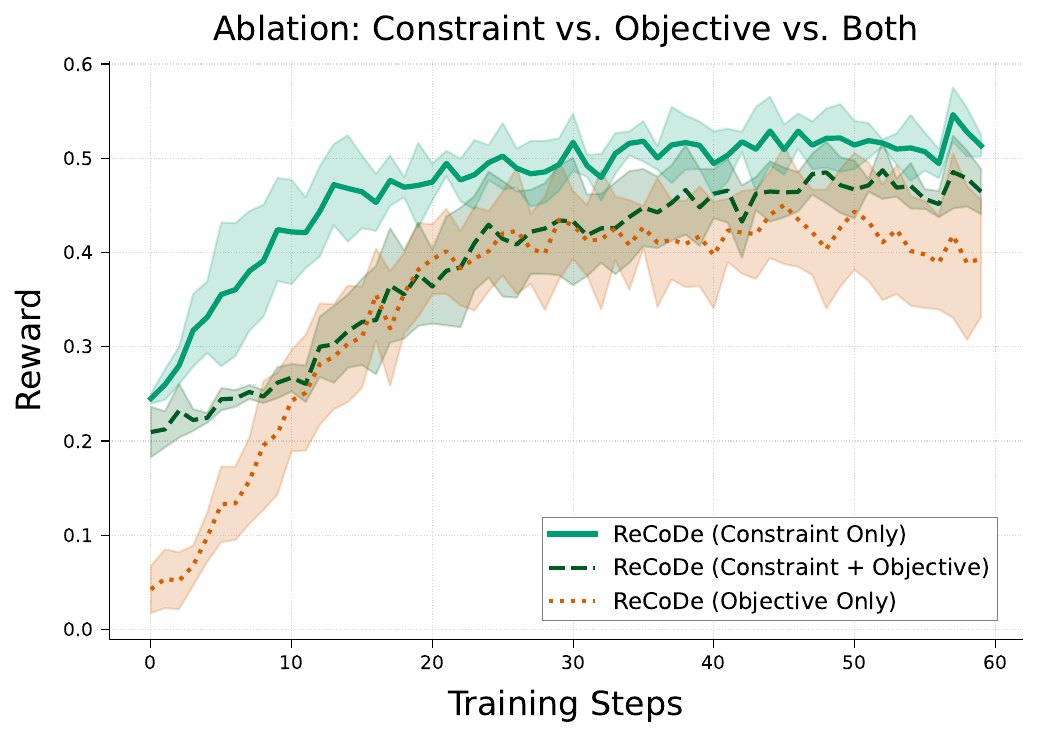}};
    \node[anchor=north west,xshift=-8pt,yshift=3pt] at (image.north west) {\textbf{(f)}};
  \end{tikzpicture}
  \phantomsubcaption
  \label{fig:constraintobjectiveablation}
\end{subfigure}

\caption{%
  \textbf{(\subref{fig:complexityvsreward})}~Complexity vs.\ Reward in Waypoint Navigation:
  ReCoDe consistently outperforms both pure MARL and the handcrafted controller across different agent radii, demonstrating robustness in both high- and low-task complexities. Shaded regions indicate standard deviations. 
  \textbf{(\subref{fig:sample_efficiency})}~Sample Efficiency:
  In Waypoint Navigation with agent radius = $0.1$, ReCoDe quickly converges to near-optimal reward, whereas pure MARL remains suboptimal even after 500 steps.
  \textbf{(\subref{fig:collision_rew})}~Collision penalties during training of the Narrow Corridor scenario are near-zero for ReCoDe (\(-0.0001\) on average), Online CBF (\(-0.06\)) and pure RL (\(-9.4\)).
  \textbf{(\subref{fig:b_vs_y_position})}--\textbf{(\subref{fig:b_vs_num_neighbors})}
  Plots of the learned constraint radius (\(b\)) vs.\ the agent’s \(y\)-position and number of neighbors at various stages of training in the Narrow Corridor scenario.  Shaded regions indicate $\frac{1}{2}$ std. 
  \textbf{(\subref{fig:constraintobjectiveablation})}~Ablation:
  Learning constraints vs. objective vs. both in the Narrow Corridor setting (4 random seeds).
}
\label{fig:spanning_figure}
\vspace{-3mm}
\end{figure*}

\headline{MARL/Handcrafted Controller Comparison.} We examined how MARL and the handcrafted controller compare to ReCoDe under easier conditions by varying each agent’s radius in \textit{Waypoint Navigation}. Smaller radii make collisions less likely, simplify navigation and require less intricate  coordination. Figure~\ref{fig:complexityvsreward} shows that even with smaller agents, ReCoDe outperforms both pure MARL and the handcrafted controller, and so might be beneficial even for less coordination-heavy scenarios. Furthermore, compared to MARL, ReCoDe demonstrates much faster training convergence: as shown in Figure~\ref{fig:sample_efficiency}, whereas pure MARL still underperforms substantially after 500 training steps of 120k frames each, ReCoDe reaches excellent performance in 20 steps. Another critical consideration is \textit{safety during training}. As shown in Figure \ref{fig:collision_rew}, ReCoDe consistently maintains near-zero collision rates during training, a key advantage of hybrid methods over end-to-end MARL.

\headline{Mechanism.} Proposition~\ref{thm:uncertaintyestimate} suggests that agents should shrink the uncertainty radius~\(b\) when the learned policy is reliable and enlarge it when the handcrafted objective offers stronger signal.  To test whether ReCoDe learns this behavior, we logged \(b\) at three training checkpoints (0.96\,M, 1.92\,M and 8.64\,M steps) in the \textit{Narrow Corridor} task  over 100 six-robot episodes with random initial states, collecting data from agents aiming for the green region. Figures~\ref{fig:b_vs_num_neighbors}–\ref{fig:b_vs_y_position} plot \(b\) against each robot’s neighbor count and \(y\)-position (robots with \(y\!>\!1.5\) have reached the goal).  Early in training \(b\) is uniformly small, indicating near-total reliance on the learned policy; as learning progresses the mean radius grows, allowing the default controller to contribute more.  We find that \(b\) correlates negatively with number of nearby agents (\(r\!\approx\!-0.03,\;p<10^{-40}\)) and positively with $y$/goal proximity (\(r\!\approx\!0.08,\;p<10^{-200}\)).  Thus ReCoDe tightens \(b\) to resolve likely deadlocks (as the learned policy is better at such coordination) and relaxes it once the path is clear (as the default controller makes more efficient, greedy movements), matching the strategy predicted by our analysis.

\headline{Ablation.} To test whether performance can be improved by learning the objective, we compared three variants of ReCoDe in \textit{Narrow Corridor}: \emph{standard} ReCoDe, which learns only the quadratic-constraint parameters; a variant that keeps this constraint fixed but lets the policy learn the \textit{goal position} of the objective; and a variant that learns both the constraint and the target position simultaneously. Although the third variant is the most expressive, the constraint-only version reached its peak sooner and achieved the highest return, while the two objective-learning variants converged more slowly and plateaued at lower scores (Fig.  \ref{fig:constraintobjectiveablation}), supporting our claim that the fixed, expert-designed objective provides a valuable inductive bias that is diluted once it becomes a moving target.


\headline{Robot Demonstration.} We transferred the policy trained in the \textit{Narrow-Corridor} simulation
directly to six holonomic ground robots~\citep{JanCam2024}.  
The physical arena matches the simulation scale: a $90\,$cm-wide, $6.4\,$m-long
corridor.  Robots exchange messages only when separated by $\le 1.5\,$m to mimic
the limited communication range used during training. In our trials, the handcrafted  controller 
dead-locked in every run, typically when two teams met near the midpoint.
With ReCoDe’s learned quadratic constraint active, all six robots consistently complete the swap without violating safety margins. These  results corroborate the simulation study: ReCoDe’s online
constraint generation resolved reciprocal blocking despite real-world noise stemming from positional tracking errors, communication delays, and actuator limitations in precisely executing intended commands.  We further experiment with learning \textit{linear}  constraints, and find this encourages the robots to avoid partial solutions, but does not affect reward--find more details in App.  \ref{appendix:differentconstraintforms}. Figure~\ref{fig:robomaster_exp} overlays the trajectories; a video is available at \videourl. 


\section{Discussion}
\vspace{-1mm}
We presented ReCoDe, a hybrid control framework that combines expert knowledge with reinforcement learning to augment constrained optimization-based controllers. Our experiments, focusing on applications of ReCoDe to navigation and consensus, show that this combination outperforms both of its components in isolation: handcrafted controllers respect safety yet coordinate poorly, and pure MARL explores freely but lacks local precision and safety guarantees. ReCoDe overcomes both weaknesses by constraining each agent in response to its neighbors’ intents, enabling adaptability when the handcrafted controller would stall, but falling back on that controller when the learned policy is uncertain. As a result it consistently outperforms other methods in our tested scenarios, requiring fewer samples than pure learning while not violating the user-defined safety constraints.

\section{Limitations}
\vspace{-1mm}
\textbf{(i)} Our experiments demonstrate the performance improvements of ReCoDe in intricate navigation tasks, and complex tasks involving both navigation and multi-objective consensus. However, due to the significant work involved in setting up relevant experimental scenarios, we have not yet studied ReCoDe in non-navigation settings such as multi-agent manipulation, leaving open the question of how it compares to existing baselines in such settings. This will be a topic for future work.

\textbf{(ii)} ReCoDe currently assumes the underlying optimization problem is convex. In principle,  ReCoDe can be extended to nonconvex optimization, but the user must provide a method of solving the optimization problem that is efficient enough to collect data for. If the user is fine with sub-optimal solutions, there are many methods (e.g., gradient descent) for solving nonconvex optimization problems, thus enabling us to extend ReCoDe to such settings. We note that because we are modifying the user-provided controller anyway, it is likely not important for the user to use a solver that finds the global optimum, and so this is not necessarily a fundamental limitation. However, as we did not explore nonconvex, sub-optimal solvers in this work, we cannot decisively comment on how well ReCoDe works in such settings.

\textbf{(iii)} Finally, data collection in our setting is computationally demanding to scale up to a great number of agents due to the need to many solve optimization problems. These optimization problems are challenging to parallelize efficiently, as most solvers use complex, branching control flows better suited for CPUs rather than GPUs, potentially limiting the scalability of our method. To address this limitation, we explored GPU-compatible solvers such as qpth \citep{amos2017optnet} and JAXOpt \citep{jaxopt_implicit_diff}. While these recently developed solvers were specifically designed to address such scalability challenges, we observed minimal computational advantages over CPU-based solvers given our relatively inexpensive RL  experiments. However, tests we performed indicate that these GPU-based solvers demonstrate significantly better scaling properties when handling larger numbers of problems, suggesting they should be preferred over CPU-based alternatives like CVXPYLayers \cite{cvxpylayers2019} for larger-scale experiments. Scaling up our framework using such GPU-compatible solvers can expand the scope of our method to larger systems of agents, and further exploration is required to find the optimal setup.

\bibliography{bibliography}

\begin{thebibliography}{53}
\providecommand{\natexlab}[1]{#1}
\providecommand{\url}[1]{\texttt{#1}}
\expandafter\ifx\csname urlstyle\endcsname\relax
  \providecommand{\doi}[1]{doi: #1}\else
  \providecommand{\doi}{doi: \begingroup \urlstyle{rm}\Url}\fi

\bibitem[Blumenkamp et~al.(2024)Blumenkamp, Shankar, Bettini, Bird, and Prorok]{JanCam2024}
J.~Blumenkamp, A.~Shankar, M.~Bettini, J.~Bird, and A.~Prorok.
\newblock \href{https://arxiv.org/abs/2405.02198}{The Cambridge RoboMaster: An Agile Multi-Robot Research Platform}.
\newblock In \emph{International Symposium on Distributed Autonomous Robotic Systems (DARS)}, 2024.

\bibitem[Alonso-Mora et~al.(2015)Alonso-Mora, Naegeli, Siegwart, and Beardsley]{alonso2015collision}
J.~Alonso-Mora, T.~Naegeli, R.~Siegwart, and P.~Beardsley.
\newblock \href{https://idp.springer.com/authorize/casa?redirect_uri=https://link.springer.com/article/10.1007/s10514-015-9429-0&casa_token=QByq7Zgi3p4AAAAA:n-56Bf5bOfGUZnr2EKghJrJc2gBrnB1cvEk1yaXZsBMWVU_LU2mxGg9xJttpjJcDOoYI39_yOmB2qyjd7g}{Collision avoidance for aerial vehicles in multi-agent scenarios}.
\newblock \emph{Autonomous Robots}, 39:\penalty0 101--121, 2015.

\bibitem[Merkt et~al.(2019)Merkt, Ivan, and Vijayakumar]{merkt2019continuous}
W.~Merkt, V.~Ivan, and S.~Vijayakumar.
\newblock \href{https://ieeexplore.ieee.org/abstract/document/8967641?casa_token=5mmHAqsKC6oAAAAA:7UUtxPMTgowJT4Ncp0N2PCasncHL_NYQVUxIgj9fBlNa9CkHfpJbSdxdl_dy2bd90HIQv_QT_gI}{Continuous-time collision avoidance for trajectory optimization in dynamic environments}.
\newblock In \emph{IEEE/RSJ International Conference on Intelligent Robots and Systems (IROS)}, 2019.

\bibitem[Gregory(2018)]{gregory2018constrained}
J.~Gregory.
\newblock \emph{\href{https://www.taylorfrancis.com/books/mono/10.1201/9781351070867/constrained-optimization-calculus-variations-optimal-control-theory-gregory}{Constrained optimization in the calculus of variations and optimal control theory}}.
\newblock Chapman and Hall/CRC, 2018.

\bibitem[Gao et~al.(2023)Gao, Yang, and Prorok]{gao2023onlinecbf}
Z.~Gao, G.~Yang, and A.~Prorok.
\newblock \href{https://ieeexplore.ieee.org/abstract/document/10416796/?casa_token=f4WRo97Ux-gAAAAA:fznIARBiiClnS4jJDLD2_kpHbq6RBN-KREJnvpbj4Vnd8rDzgUkCU7MrwePw_gGOdBRTskjRVPU}{Online control barrier functions for decentralized multi-agent navigation}.
\newblock In \emph{IEEE International Symposium on Multi-Robot and Multi-Agent Systems (MRS)}, 2023.

\bibitem[Busoniu et~al.(2008)Busoniu, Babuska, and De~Schutter]{busoniu2008comprehensive}
L.~Busoniu, R.~Babuska, and B.~De~Schutter.
\newblock \href{https://ieeexplore.ieee.org/abstract/document/4445757/?casa_token=H22XaTCkmEsAAAAA:4b9W9gXNjq0ABFmDmTojKdAbeL6qFTB2M9IHa7w3wO5SI8sma0WE5guSXB55CkNrXMIJJN7QElo}{A comprehensive survey of multiagent reinforcement learning}.
\newblock \emph{IEEE Transactions on Systems, Man, and Cybernetics, Part C (Applications and Reviews)}, 38\penalty0 (2):\penalty0 156--172, 2008.

\bibitem[Canese et~al.(2021)Canese, Cardarilli, Di~Nunzio, Fazzolari, Giardino, Re, and Span{\`o}]{canese2021multi}
L.~Canese, G.~C. Cardarilli, L.~Di~Nunzio, R.~Fazzolari, D.~Giardino, M.~Re, and S.~Span{\`o}.
\newblock \href{https://www.mdpi.com/2076-3417/11/11/4948}{Multi-agent reinforcement learning: A review of challenges and applications}.
\newblock \emph{Applied Sciences}, 11\penalty0 (11):\penalty0 4948, 2021.

\bibitem[Gronauer and Diepold(2022)]{gronauer2022multi}
S.~Gronauer and K.~Diepold.
\newblock \href{https://link.springer.com/article/10.1007/s10462-021-09996-w}{Multi-agent deep reinforcement learning: a survey}.
\newblock \emph{Artificial Intelligence Review}, 55\penalty0 (2):\penalty0 895--943, 2022.

\bibitem[Alshiekh et~al.(2018)Alshiekh, Bloem, Ehlers, K{\"o}nighofer, Niekum, and Topcu]{alshiekh2018safeshielding}
M.~Alshiekh, R.~Bloem, R.~Ehlers, B.~K{\"o}nighofer, S.~Niekum, and U.~Topcu.
\newblock \href{https://ojs.aaai.org/index.php/AAAI/article/view/11797}{Safe reinforcement learning via shielding}.
\newblock In \emph{Proceedings of the AAAI Conference on Artificial Intelligence}, volume~32, 2018.

\bibitem[R. et~al.(2024)R., A., S., and S.]{romero2024actorcriticmodelpredictivecontrol}
A.~R., E.~A., Y.~S., and D.~S.
\newblock \href{https://arxiv.org/abs/2306.09852}{Actor-Critic Model Predictive Control: Differentiable Optimization meets Reinforcement Learning}, 2024.

\bibitem[Van~den Berg et~al.(2008)Van~den Berg, Lin, and Manocha]{van2008reciprocalvelocityobstaclesRVO}
J.~Van~den Berg, M.~Lin, and D.~Manocha.
\newblock Reciprocal velocity obstacles for real-time multi-agent navigation.
\newblock In \emph{2008 IEEE international conference on robotics and automation}, pages 1928--1935. Ieee, 2008.

\bibitem[Earl and D'Andrea(2002)]{earl2002modeling}
M.~G. Earl and R.~D'Andrea.
\newblock \href{https://ieeexplore.ieee.org/abstract/document/1184476/?casa_token=6mw4gCPmmPAAAAAA:lY7ZMJMmOF9ibXLNjkHMrAQC50JAPBrwX4pQoqQvPzGXrAxYviOdkuYYfmR-wEf_-xfkaHcK_t4}{Modeling and control of a multi-agent system using mixed integer linear programming}.
\newblock In \emph{IEEE Conference on Decision and Control (CDC)}, 2002.

\bibitem[Fallahi et~al.(2019)Fallahi, Rosenberger, Chen, Lee, and Wang]{fallahi2019linear}
A.~Fallahi, J.~M. Rosenberger, V.~C.~P. Chen, W.~Lee, and S.~Wang.
\newblock \href{https://ieeexplore.ieee.org/abstract/document/8932474/}{Linear programming for multi-agent demand response}.
\newblock \emph{IEEE Access}, 7:\penalty0 181479--181490, 2019.

\bibitem[Nocedal and Wright(2006)]{nocedal2006quadratic}
J.~Nocedal and S.~J. Wright.
\newblock \href{https://link.springer.com/content/pdf/10.1007/978-0-387-40065-5_16.pdf}{Quadratic programming}.
\newblock \emph{Numerical Optimization}, pages 448--492, 2006.

\bibitem[Nguyen and Sreenath(2016)]{nguyen2016exponential}
Q.~Nguyen and K.~Sreenath.
\newblock \href{https://ieeexplore.ieee.org/abstract/document/7524935?casa_token=v6uqNunoT1sAAAAA:hOKrzLQY7dMN0ocV_fvKNpY3aNTDmQ5LaMTSm1dFnKibptQjGkNhqZ6zmLfal20k3x8ClZuUS-A}{Exponential control barrier functions for enforcing high relative-degree safety-critical constraints}.
\newblock In \emph{IEEE American Control Conference (ACC)}, 2016.

\bibitem[Tran et~al.(2017)Tran, Prodan, and Lef{\`e}vre]{tran2017nonlinear}
N.~Q.~H. Tran, I.~Prodan, and L.~Lef{\`e}vre.
\newblock \href{https://ieeexplore.ieee.org/abstract/document/8107082/?casa_token=yjrrg2v4-4sAAAAA:j8iNCBMZWYAUwHnP3WSAUhlv45l834MJzw1CnaZ8nJxVygjYW6dL9mXEv7gX04ed57dqVgClCbE}{Nonlinear optimization for multi-agent motion planning in a multi-obstacle environment}.
\newblock In \emph{IEEE International Conference on System Theory, Control and Computing (ICSTCC)}, 2017.

\bibitem[Chu et~al.(2019)Chu, Wang, Codec{\`a}, and Li]{chu2019multi}
T.~Chu, J.~Wang, L.~Codec{\`a}, and Z.~Li.
\newblock \href{https://ieeexplore.ieee.org/abstract/document/8667868/?casa_token=awQNHiBCDLYAAAAA:ee-3z5cfTHn3xyHje21Ivq4cn7At7Nn55PcQBfK4r8a9TTPnBFb_OsKiexOcLm2G1UOmErY6Jp0}{Multi-agent deep reinforcement learning for large-scale traffic signal control}.
\newblock \emph{IEEE Transactions on Intelligent Transportation Systems}, 21\penalty0 (3):\penalty0 1086--1095, 2019.

\bibitem[Xue and Chen(2023)]{xue2023multi}
Y.~Xue and W.~Chen.
\newblock \href{https://ieeexplore.ieee.org/abstract/document/10192519?casa_token=4maw1gi05TsAAAAA:k3gYERN4A9-z1b5CUghB82xKgTV8yQ863wequeAa9SoTVBeTwmsDiYNJ8SA8gryQaIOSmIMMHwA}{Multi-agent deep reinforcement learning for UAVs navigation in unknown complex environment}.
\newblock \emph{IEEE Transactions on Intelligent Vehicles}, 2023.

\bibitem[Ning and Xie(2024)]{ning2024survey}
Z.~Ning and L.~Xie.
\newblock \href{https://www.researchgate.net/publication/378311162_A_survey_on_multi-agent_reinforcement_learning_and_its_application}{A survey on multi-agent reinforcement learning and its application}.
\newblock \emph{Journal of Automation and Intelligence}, 2024.

\bibitem[Amato(2024)]{amato2024introduction}
C.~Amato.
\newblock \href{https://arxiv.org/abs/2409.03052}{An introduction to centralized training for decentralized execution in cooperative multi-agent reinforcement learning}.
\newblock \emph{arXiv preprint arXiv:2409.03052}, 2024.

\bibitem[Fiacco(1983)]{fiacco1983introduction}
A.~V. Fiacco.
\newblock \href{https://www.sidalc.net/search/Record/cat-unco-ar-34375/Description}{Introduction to sensitivity and stability analysis in non linear programming}.
\newblock 1983.

\bibitem[Lobo et~al.(1998)Lobo, Vandenberghe, Boyd, and Lebret]{lobo1998applications}
M.~S. Lobo, L.~Vandenberghe, S.~Boyd, and H.~Lebret.
\newblock \href{https://www.sciencedirect.com/science/article/pii/S0024379598100320}{Applications of second-order cone programming}.
\newblock \emph{Linear algebra and its applications}, 284\penalty0 (1-3):\penalty0 193--228, 1998.

\bibitem[Nayak et~al.(2023)Nayak, Choi, Ding, Dolan, Gopalakrishnan, and Balakrishnan]{nayak2023scalable}
S.~Nayak, K.~Choi, W.~Ding, S.~Dolan, K.~Gopalakrishnan, and H.~Balakrishnan.
\newblock Scalable multi-agent reinforcement learning through intelligent information aggregation.
\newblock In \emph{International Conference on Machine Learning}, pages 25817--25833. PMLR, 2023.

\bibitem[Amos and Kolter(2017)]{amos2017optnet}
B.~Amos and J.~Z. Kolter.
\newblock \href{https://proceedings.mlr.press/v70/amos17a.html}{Optnet: Differentiable optimization as a layer in neural networks}.
\newblock In \emph{International Conference on Machine Learning (ICML)}, 2017.

\bibitem[Blondel et~al.(2022)Blondel, Berthet, Cuturi, Frostig, Hoyer, Llinares-L{\'o}pez, and Vert]{jaxopt_implicit_diff}
M.~Blondel, Q.~Berthet, M.~Cuturi, R.~Frostig, S.~Hoyer, F.~Llinares-L{\'o}pez, F.and~Pedregosa, and J.~Vert.
\newblock \href{https://proceedings.neurips.cc/paper_files/paper/2022/hash/228b9279ecf9bbafe582406850c57115-Abstract-Conference.html}{Efficient and Modular Implicit Differentiation}.
\newblock In \emph{Advances in Neural Information Processing Systems (NeurIPS)}, 2022.

\bibitem[Agrawal et~al.(2019)Agrawal, Amos, Barratt, Boyd, Diamond, and Kolter]{cvxpylayers2019}
A.~Agrawal, B.~Amos, S.~Barratt, S.~Boyd, S.~Diamond, and Z.~Kolter.
\newblock \href{https://proceedings.neurips.cc/paper/2019/hash/9ce3c52fc54362e22053399d3181c638-Abstract.html}{Differentiable Convex Optimization Layers}.
\newblock In \emph{Advances in Neural Information Processing Systems (NeurIPS)}, 2019.

\bibitem[Dong(2006)]{dong2006methods}
S.~Dong.
\newblock \href{https://www.researchgate.net/profile/Shuonan-Dong-2/publication/255602767_Methods_for_Constrained_Optimization/links/00b7d53c5c41574549000000/Methods-for-Constrained-Optimization.pdf} methods for constrained optimization.
\newblock \emph{Massachusetts Institute of Technology}, 2006.

\bibitem[Zhang et~al.(2024)Zhang, Lin, Ding, and Xing]{zhang2024linear}
J.~Zhang, F.~Lin, S.~Ding, and W.~Xing.
\newblock \href{https://ieeexplore.ieee.org/abstract/document/10759563/}{Linear Programming-Based Consensus of Positive Continuous-Time Multi-Agent Systems}.
\newblock \emph{IEEE/CAA Journal of Automatica Sinica}, 11\penalty0 (12):\penalty0 2519--2521, 2024.

\bibitem[Motee and Jadbabaie(2009)]{motee2009distributed}
N.~Motee and A.~Jadbabaie.
\newblock \href{https://ieeexplore.ieee.org/abstract/document/5272314/?casa_token=HXbzg7P9Yn4AAAAA:CoripRm5iaBcSY-6JANxgbKKSDdEKh6yEdkUF-qPbPbtoSSWNc8Eo7fIFsjJEkwsg0FQVHO4KfY}{Distributed multi-parametric quadratic programming}.
\newblock \emph{IEEE Transactions on Automatic Control}, 54\penalty0 (10):\penalty0 2279--2289, 2009.

\bibitem[Endo et~al.(2019)Endo, Ibuki, and Sampei]{endo2019collision}
M.~Endo, T.~Ibuki, and M.~Sampei.
\newblock \href{https://ieeexplore.ieee.org/abstract/document/8814603?casa_token=Y7DiwMtZ97oAAAAA:ioQEuv8HJUGXqKRABNN9WYs0KkvMq4DiqhJcA6W5Ha5oimTvZ79VAvqY0Fn_kdzpOf5rYSqwIjQ}{Collision-free formation control for quadrotor networks based on distributed quadratic programs}.
\newblock In \emph{IEEE American Control Conference (ACC)}, 2019.

\bibitem[Romero et~al.(2024)Romero, Song, and Scaramuzza]{romero2024actor}
A.~Romero, Y.~Song, and D.~Scaramuzza.
\newblock \href{wlgyMBy8bfCiZTc}{Actor-critic model predictive control}.
\newblock In \emph{IEEE International Conference on Robotics and Automation (ICRA)}, 2024.

\bibitem[Sun and Cassandras(2016)]{sun2016optimal}
X.~Sun and C.~G. Cassandras.
\newblock \href{https://www.sciencedirect.com/science/article/pii/S0005109816302965?casa_token=qnyK6b7TCuQAAAAA:tDXMyTvJvagJzO9z1YKwNYuyFH-wKzESxCZC9GNS12mTByKMxRNUT0LLJYQ2QGONzird8Y1VYCI}{Optimal dynamic formation control of multi-agent systems in constrained environments}.
\newblock \emph{Automatica}, 73:\penalty0 169--179, 2016.

\bibitem[Chai and Hodgins(2007)]{chai2007constraint}
J.~Chai and J.~K. Hodgins.
\newblock \href{https://dl.acm.org/doi/abs/10.1145/1275808.1276387?casa_token=1i4qnhSDl-8AAAAA:Gbl2tColOyBGZaSiMlj5dZGWoQ62NWEagdhbdwnwWWUZJdDzjr7h118tk7uhHzX2JIA2oxzVkDY_7A}{Constraint-based motion optimization using a statistical dynamic model}.
\newblock In \emph{ACM SIGGRAPH papers}, pages 8--es. 2007.

\bibitem[Lin et~al.(2018)Lin, Somani, Hu, Rickert, and Knoll]{lin2018efficient}
J.~Lin, N.~Somani, B.~Hu, M.~Rickert, and A.~Knoll.
\newblock \href{https://ieeexplore.ieee.org/abstract/document/8593577/?casa_token=i9q_rESOCbEAAAAA:n4Gyyz3FSFKp5JcFoL-2mwAKk-f9pfiThefmJKBrKels3D0nijyUi43w9dLlECZgNj9D_yc5F2Q}{An efficient and time-optimal trajectory generation approach for waypoints under kinematic constraints and error bounds}.
\newblock In \emph{IEEE/RSJ International Conference on Intelligent Robots and Systems (IROS)}, 2018.

\bibitem[Mu{\~n}oyerro et~al.(2023)Mu{\~n}oyerro, Hern{\'a}ndez, Urizar, and Altuzarra]{munoyerro2023general}
A.~Mu{\~n}oyerro, A.~Hern{\'a}ndez, M.~Urizar, and O.~Altuzarra.
\newblock \href{https://journals.sagepub.com/doi/abs/10.1177/09544062221147829?casa_token=ww9bfRQebRkAAAAA:QNcQEJODDIXpqtpWjLv19kz4hvp1mZzgGVKdvlvi7HmF7hU5d1s5N72ZW0epbXEe2MbbMsIocE2EtA&casa_token=OREz-O5c6gIAAAAA:LJyvOMS3EotWsPvwJOShEkegPenkCtvbuhR7K4ZULdQIBwi02uZpVsVBsJVAZVlV8Br_G0hToybXeA}{A general automatic method for mechanism optimization based on kinematic constraints and analytical Jacobian matrix}.
\newblock \emph{Proceedings of the Institution of Mechanical Engineers, Part C: Journal of Mechanical Engineering Science}, 237\penalty0 (14):\penalty0 3181--3197, 2023.

\bibitem[Gonz{\'a}lez-Briones et~al.(2018)Gonz{\'a}lez-Briones, De~La~Prieta, Mohamad, Omatu, and Corchado]{gonzalez2018multi}
A.~Gonz{\'a}lez-Briones, F.~De~La~Prieta, M.~S. Mohamad, S.~Omatu, and J.~M. Corchado.
\newblock \href{https://www.mdpi.com/1996-1073/11/8/1928}{Multi-agent systems applications in energy optimization problems: A state-of-the-art review}.
\newblock \emph{Energies}, 11\penalty0 (8):\penalty0 1928, 2018.

\bibitem[Gao and Prorok(2023{\natexlab{a}})]{gao2023environment}
Z.~Gao and A.~Prorok.
\newblock \href{https://ieeexplore.ieee.org/abstract/document/10160813/?casa_token=xoBQeJhZJrsAAAAA:trJujR1qxjafFbMfUYhZDveLVpWep_b-KxGvJqyWOxS3FMko2VWRs70MMxb__rp9CnzAnyfwr7Q}{Environment optimization for multi-agent navigation}.
\newblock In \emph{IEEE International Conference on Robotics and Automation (ICRA)}, 2023{\natexlab{a}}.

\bibitem[Gao and Prorok(2023{\natexlab{b}})]{gao2023constrained}
Z.~Gao and A.~Prorok.
\newblock \href{https://ieeexplore.ieee.org/abstract/document/10251921/}{Constrained environment optimization for prioritized multi-agent navigation}.
\newblock \emph{IEEE Open Journal of Control Systems}, 2023{\natexlab{b}}.

\bibitem[Kornienko et~al.(2004)Kornienko, Kornienko, and Priese]{kornienko2004application}
S.~Kornienko, O.~Kornienko, and J.~Priese.
\newblock \href{https://www.sciencedirect.com/science/article/pii/S0166361504000053?casa_token=doaR2R02r28AAAAA:mBWfsChi_6LtI7s_HePjl-KfyBfXBQtWDrLSr7HSFoDcbIQCSQCQYPPWOr5IFzIyNwIVO6RilTI}{Application of multi-agent planning to the assignment problem}.
\newblock \emph{Computers in Industry}, 54\penalty0 (3):\penalty0 273--290, 2004.

\bibitem[Nedic et~al.(2010)Nedic, Ozdaglar, and Parrilo]{nedic2010constrained}
A.~Nedic, A.~Ozdaglar, and P.~A. Parrilo.
\newblock \href{https://ieeexplore.ieee.org/abstract/document/5404774/?casa_token=UR9_mBuPJhsAAAAA:hgUWc7S3bOfGB_xqYL1D5MgHa729wx2ONt2glv2MtY9ZuFy2tDeGiFzLOA0wkFhFEHupskEQc2k}{Constrained consensus and optimization in multi-agent networks}.
\newblock \emph{IEEE Transactions on Automatic Control}, 55\penalty0 (4):\penalty0 922--938, 2010.

\bibitem[Zheng and Wang(2015)]{zheng2015multi}
X.~Zheng and L.~Wang.
\newblock \href{https://www.sciencedirect.com/science/article/pii/S0957417415002390?casa_token=B_XY--pNoLMAAAAA:-YOqdIWrYYWnMKO1ubXFJUn31xIAqugPKYtsNxruBhiPkUIHu1-s_hfsAfhxz5XR4ukY_Vk00_8}{A multi-agent optimization algorithm for resource constrained project scheduling problem}.
\newblock \emph{Expert Systems with Applications}, 42\penalty0 (15-16):\penalty0 6039--6049, 2015.

\bibitem[Bu{\c{s}}oniu et~al.(2010)Bu{\c{s}}oniu, Babu{\v{s}}ka, and De~Schutter]{bucsoniu2010multi}
L.~Bu{\c{s}}oniu, R.~Babu{\v{s}}ka, and B.~De~Schutter.
\newblock \href{https://link.springer.com/chapter/10.1007/978-3-642-14435-6_7}{Multi-agent reinforcement learning: An overview}.
\newblock \emph{Innovations in Multi-Agent Systems and Applications-1}, pages 183--221, 2010.

\bibitem[Gao et~al.(2024)Gao, Yang, and Prorok]{gao2024co}
Z.~Gao, G.~Yang, and A.~Prorok.
\newblock \href{https://arxiv.org/abs/2403.14583}{Co-Optimization of Environment and Policies for Decentralized Multi-Agent Navigation}.
\newblock \emph{arXiv preprint arXiv:2403.14583}, 2024.

\bibitem[Damadam et~al.(2022)Damadam, Zourbakhsh, Javidan, and Faroughi]{damadam2022intelligent}
S.~Damadam, M.~Zourbakhsh, R.~Javidan, and A.~Faroughi.
\newblock \href{https://www.mdpi.com/2624-6511/5/4/66}{An intelligent IoT based traffic light management system: deep reinforcement learning}.
\newblock \emph{Smart Cities}, 5\penalty0 (4):\penalty0 1293--1311, 2022.

\bibitem[Pan et~al.(2022)Pan, Liu, Zhong, Yang, Zhu, and Wang]{pan2022mate}
X.~Pan, M.~Liu, F.~Zhong, Y.~Yang, S.-C. Zhu, and Y.~Wang.
\newblock \href{https://proceedings.neurips.cc/paper_files/paper/2022/hash/b2a1c152f14a4b842a9ddb3bd84c62a1-Abstract-Datasets_and_Benchmarks.html}{Mate: Benchmarking multi-agent reinforcement learning in distributed target coverage control}.
\newblock In \emph{Advances in Neural Information Processing Systems (NeurIPS)}, 2022.

\bibitem[Aydemir and Cetin(2023)]{aydemir2023multi}
F.~Aydemir and A.~Cetin.
\newblock \href{https://avesis.gazi.edu.tr/yayin/8b490f01-0890-4b0e-8870-74f43ba3925d/multi-agent-dynamic-area-coverage-based-on-reinforcement-learning-with-connected-agents}{Multi-agent dynamic area coverage based on reinforcement learning with connected agents}.
\newblock \emph{Computer Systems Science and Engineering}, 45\penalty0 (1), 2023.

\bibitem[De~Witt et~al.(2020)De~Witt, Gupta, Makoviichuk, Makoviychuk, Torr, Sun, and Whiteson]{de2020independent}
C.~S. De~Witt, T.~Gupta, D.~Makoviichuk, V.~Makoviychuk, P.~H.~S. Torr, M.~Sun, and S.~Whiteson.
\newblock \href{https://arxiv.org/abs/2011.09533}{Is independent learning all you need in the starcraft multi-agent challenge?}
\newblock \emph{arXiv preprint arXiv:2011.09533}, 2020.

\bibitem[Schulman et~al.(2017)Schulman, Wolski, Dhariwal, Radford, and Klimov]{schulman2017proximal}
J.~Schulman, F.~Wolski, P.~Dhariwal, A.~Radford, and O.~Klimov.
\newblock \href{https://arxiv.org/abs/1707.06347}{Proximal policy optimization algorithms}.
\newblock \emph{arXiv preprint arXiv:1707.06347}, 2017.

\bibitem[Brody et~al.(2022)Brody, Alon, and Yahav]{brody2021attentive}
S.~Brody, U.~Alon, and E.~Yahav.
\newblock \href{https://openreview.net/forum?id=F72ximsx7C1}{How attentive are graph attention networks?}
\newblock In \emph{International Conference on Learning Representations (ICLR)}, 2022.

\bibitem[Bettini et~al.(2022)Bettini, Kortvelesy, Blumenkamp, and Prorok]{bettini2022vmas}
M.~Bettini, R.~Kortvelesy, J.~Blumenkamp, and A.~Prorok.
\newblock \href{https://link.springer.com/chapter/10.1007/978-3-031-51497-5_4}{VMAS: A Vectorized Multi-Agent Simulator for Collective Robot Learning}.
\newblock \emph{International Symposium on Distributed Autonomous Robotic Systems (DARS)}, 2022.

\bibitem[Bettini et~al.(2024)Bettini, Prorok, and Moens]{bettini2024benchmarl}
M.~Bettini, A.~Prorok, and V.~Moens.
\newblock Benchmarl: Benchmarking multi-agent reinforcement learning.
\newblock \emph{Journal of Machine Learning Research}, 25\penalty0 (217):\penalty0 1--10, 2024.

\bibitem[Blumenkamp et~al.(2024)Blumenkamp, Shankar, Bettini, Bird, and Prorok]{blumenkamp2024cambridge}
J.~Blumenkamp, A.~Shankar, M.~Bettini, J.~Bird, and A.~Prorok.
\newblock \href{https://arxiv.org/abs/2405.02198}{The Cambridge RoboMaster: An Agile Multi-Robot Research Platform}.
\newblock In \emph{IEEE International Symposium on Distributed Robotic Systems (DARS)}, 2024.

\bibitem[Shankar et~al.(2021)Shankar, Elbaum, and Detweiler]{shankar2021freyja}
A.~Shankar, S.~Elbaum, and C.~Detweiler.
\newblock \href{https://ieeexplore.ieee.org/abstract/document/9562076?casa_token=i--jG4JnBm8AAAAA:QWFuOWOrtgw0YXbKNunhVdQfx_eMv3DrnMz6uR2K7myqfujd2KP2myz2t_2VlebUOC_bigvlx_8}{Freyja: A full multirotor system for agile \& precise outdoor flights}.
\newblock In \emph{IEEE International Conference on Robotics and Automation (ICRA)}, 2021.

\end{thebibliography}
\clearpage

\appendix
\section*{Appendix}
\section{Further Related Work on Constrained Optimization and Multi-Agent RL}
\label{appendix:extendedrelatedworks}

\textbf{Constrained Optimization:} Constrained optimization is one of the main tools to address multi-agent motion control problems with system dynamics in the continuous domain. It provides a structured methodology for choosing agent actions, where the objective function for optimization encodes the mission goal and the constraints adhere to certain operational, physical or environmental constraints \citep{gregory2018constrained}. Depending on the problem’s complexity, various algorithms have been used in the literature \cite{dong2006methods}.  When objective functions and constraints are linear, linear programming efficiently coordinates multi-agent movements \citep{earl2002modeling, fallahi2019linear, zhang2024linear}. For quadratic objectives (e.g., minimizing energy or path deviation), quadratic programming (QP) and Second-order Cone Programming (SOCP) are widely used \citep{nocedal2006quadratic, motee2009distributed, fiacco1983introduction}. They are often combined with control barrier functions (CBFs) and control Lyapunov functions (CLFs) for collision avoidance and target tracking \citep{nguyen2016exponential, gao2023onlinecbf}, and find applications in formation control \citep{endo2019collision} and model-predictive control \citep{romero2024actor}. These methods integrate system dynamics and constraints to provide solutions in convex settings. When dealing with complex dynamics, nonlinear programming remains a possibility \citep{tran2017nonlinear, sun2016optimal}, at higher computational cost.

\headline{Constraints:} In (multi-)agent control, constrained optimization problems generally consider four types of constraints: (i) collision avoidance constraints; (ii) kinematic constraints; (iii) environmental constraints and (iv) task-specific constraints. In particular, collision avoidance is critical for multi-agent motion control to ensure safe and efficient operation in shared spaces, which is a fundamental requirement in many existing works \cite{alonso2015collision, merkt2019continuous}. Kinematic constraints describe limitations on agent movements given their physical configuration and properties, which is important to ensure physically feasible trajectories for multi-agent systems and reduces sim-to-real gap compared to traditional discrete multi-agent path finding methods \cite{chai2007constraint, lin2018efficient, munoyerro2023general}. Similarly, environmental constraints impose restrictions on multi-agent systems based on characteristics and conditions of the environment in which multi-agent systems operate, including physical constraints such as walls or obstacles and operational constraints such as energy limitations. These constraints play a critical role in ensuring safe, efficient and robust operations of multi-agent systems in real-world environments \cite{gonzalez2018multi, gao2023environment, gao2023constrained}. Finally, task-specific constraints arise from specific objectives or missions that agents need to accomplish, which are fundamental to align motion planning and control strategies with functional goals of the system, such as goal constraints, time constraints and assignment constraints \cite{kornienko2004application, nedic2010constrained, zheng2015multi}. These constraints must be carefully designed to avoid conflicting with each other, which could make the optimization problem infeasible.

\textbf{Learning-based methods} for multi-agent control have attracted significant attention in recent years  \cite{gronauer2022multi}. These methods parameterize the control policies of multi-agent systems with neural networks and train their parameters with gradient-based algorithms, among which multi-agent reinforcement learning (MARL) is highly notable. In MARL, agents learn to make decisions by interacting with the environment and optimizing their policies based on environmental feedback and communication. 
Agents can be cooperative, competitive or a mix of both \cite{busoniu2008comprehensive, bucsoniu2010multi, canese2021multi}. MARL has been applied to multi-agent navigation \cite{xue2023multi, gao2024co}, traffic management \cite{chu2019multi, damadam2022intelligent}, coverage control \cite{pan2022mate, aydemir2023multi}, among other domains. Learning-based methods do not need to solve optimization problems and are computationally efficient. However, they lose theoretical guarantees compared to optimization-based methods and do not always converge to a good policy.

\section{Implementation Details}
\label{appendix:implementation-details}

 In our setting, we train agents using Multi-Agent Proximal Policy Optimization (MAPPO)~\citep{de2020independent}, a multi-agent variant of PPO \citep{schulman2017proximal} where each agent optimizes its policy using shared experience from the environment while maintaining a centralized value function to improve coordination and stability during training. We find that IPPO--a simpler, more decentralized multi-agent learning version of PPO--also works well.

 Both our agents' actor policy and critic policy consist of a Graph Attention Network v2 (\texttt{GATv2Conv})~\citep{brody2021attentive} layer that processes agent observations as node and edge features (e.g., node features might be the agent's goal point, and edge features might consist of relative positions and velocities). The \texttt{GATv2Conv} layer computes attention coefficients between connected agents to dynamically weigh the importance of neighboring information, producing updated node. The output of the GNN is decoded by an MLP with one hidden layer (128 units, \texttt{Tanh} activation) that maps the GNN embeddings to constraint parameters \(\boldsymbol{\theta}_i(t)\).

To solve optimization problems we use CVXPYLayers \citep{cvxpylayers2019}, which has some built-in support for parallelization that enables us to solve large batches of problems faster. We use VMAS \citep{bettini2022vmas} as the backbone of our vectorized multi-agent environment.

We find that ReCoDe's performance is not particularly sensitive to choice of hyperparameters and use the defaults of BenchMARL \cite{bettini2024benchmarl} to set the parameters of MAPPO and the GNN layers in all experiments. 


\section{Benefits of the Quadratic Constraint}
\label{appendix:whynotlinearconstraints}

When considering what type of constraint to augment the handcrafted controller with, one might be tempted to use general linear constraints, expressed as $\mathbf{a}_i^\top \mathbf{u}_i(t) \leq b_i$, where $\mathbf{a}_i \in \mathbb{R}^m$ and $b_i \in \mathbb{R}$ are parameters that the agent can adjust, as these are the simplest and most efficient types of constraints to solve for. While these constraints are beneficial in specific scenarios, and we do find them useful in our ground robot demonstration (see Appendix \ref{appendix:differentconstraintforms}), they have an important drawback: if an agent is allowed to parametrize $k$ such constraints, the total number of parameters becomes $k(m + 1)$, growing quickly with $k$ and $m$. However, linear constraints have a serious drawback: a small value of $k$ severely limits the agent's ability to control the outcome of its optimization problem. Specifically, with fewer than $m+1$ linear constraints, it is impossible to specify a bounded region in $\mathbb{R}^m$ that contains an $\epsilon$-ball around a point, which is necessary for fine-grained control. This suggests setting $k \geq m+1$, making the dimension of our action space at least $(m+1)^2$. However, such an action space scales poorly with $m$, requiring more samples for the agent to explore and learn effective policies, and making it difficult to generalize from limited data. We conclude that this choice of constraints is suboptimal.

We instead augment the handcrafted controller with a single quadratic constraint $\| \mathbf{u}_i(t) - \mathbf{a}_i(t) \|_2 \leq b_i(t) + s$ defining the center of a ball and its radius, as described in the paper. The action space dimension of this constraint is $\dim(\theta_i) = m + 1$--significantly smaller than $(m + 1)^2$ when using multiple linear constraints. By adjusting $\mathbf{a}_i(t)$ and $b_i(t)$, the agent can position the quadratic constraint's feasible region anywhere in the control input space and adjust its size, allowing for a wide range of control actions. This makes the quadratic constraint a flexible choice. 

In spite of the above, there are some situations where we find that learning linear constraints sometimes produces desirable behavior. Please see Appendix \ref{appendix:differentconstraintforms} for details.

\section{Proof of Proposition \ref{thm:trajectory_tracking}}
\label{appendix:proofoftrajectorysafety}

\begin{proof}
Consider a feasible desired trajectory $\{(x^*(t), u^*(t))\}_{t=1}^T$. To prove the claim, we must show that we can choose $(a(t), b(t))$ and $\lambda_0$ so that the optimal $u^{\text{opt}}(t)$ never strays more than $\varepsilon$ from $u^*(t)$. The main subtlety is that $u^{\text{opt}}(t)$ affects $x(t+1)$ and thus $\mathbf{o}_i(t+1)$, potentially changing feasibility at future steps. We handle this by leveraging the continuity of the dynamics and observations. We prove the claim by induction, assuming it holds for any $\varepsilon > 0$ up to time $t$, and showing it also holds at time $t + 1$.

First we show the base case $t = 1$. The agent's state is $x^*(1)$, trivially satisfying $\|x(1) - x^*(1)\| = 0 \leq \varepsilon$. Choose $a(1)=u^*(1)$ and $b(1)=\varepsilon$. This introduces the constraint $\|u^{\text{opt}}(1)-u^*(1)\|\leq \varepsilon + s_0$ to the optimization problem (\ref{eq:final_optimization_problem}), where $s_0$ is a slack variable. Note that as $\lambda_0 \to \infty$, any positive slack $s_0>0$ becomes increasingly costly for the optimization problem. For large enough $\lambda_0$, if a solution with $s_0=0$ exists, the solver will prefer it over any with $s_0>0$. Thus, 
by assumption, $u^*(1)$ is feasible (it lies within $\mathcal{U}_i^0(\mathbf{o}_i(1))$), and there exists a sufficiently large $\lambda_0$ for which our construction will ensure $s_0=0$, implying that $u^{\text{opt}}(1)$ must be within the $\varepsilon$-ball around $u^*(1)$ to remain optimal. Hence, $\|u^{\text{opt}}(1)-u^*(1)\|\leq \varepsilon$. This establishes the inductive claim for $t=1$.

Now assume the induction holds up to $t$, and we shall show it holds for $t+1 \leq T$. By assumption, for any $\varepsilon_0$, there exists a sequence of quadratic constraint parameters that results in our agent having state $\|x(t) - x^*(t)\| < \varepsilon_0$ at time $t$. Moreover, there is a quadratic  constraint such that the unique optimal solution \(u^{\text{opt}}(t)\) of the optimization problem \eqref{eq:final_optimization_problem} satisfies $\|u^{\text{opt}}(t) - u^*(t)\| \leq \varepsilon_0$. We know executing action $u^*(t)$ from state $x^*(t)$ results in our agent having state $x^*(t+1)$. By continuity, for any $\epsilon_1 > 0$ there exists small enough $\varepsilon_0$ such that executing $u^{\text{opt}}(t)$ results in our agent having state $x(t+1)$ satisfying $\|x(t+1)-x^*(t+1)\| < \varepsilon_1$. By assumption, $u^*(t+1)$ is strictly feasible when our agent's state is $x^*(t+1)$. By our continuity assumptions regarding \eqref{eq:final_optimization_problem}, we may choose $\varepsilon_1 > 0$ to be small enough such that $u^*(t+1)$ is strictly feasible when our agent's state is $x(t+1)$. Given such a $\varepsilon_1$, as in the base case, we can set $a(t+1)=u^*(t+1)$ and $b(t+1)=\varepsilon$  and fix $\lambda_0$ large enough to ensure $\|u^{\text{opt}}(t+1)-u^*(t+1)\|\leq \varepsilon$. This establishes the inductive claim for $t+1$.
\end{proof}

\section{Proof of Proposition \ref{thm:uncertaintyestimate}}
\label{appendix:proofofuncertaintyestimate}

\begin{proof}
By the directional–derivative assumption, integrating along
$x\mapsto a(\mathbf{o})+x\mathbf d$ for $x\in[0,r]$ gives
\(
   -J_i\bigl(\mathbf{o},a(\mathbf{o})+r\mathbf d\bigr)
   \;\ge\;
   -J_i\bigl(\mathbf{o},a(\mathbf{o})\bigr)+r\delta_2.
\)
A sufficiently large $\lambda_0$ forces
$u^{\mathrm{opt}}\in B_{r}\!\bigl(a(\mathbf{o})\bigr)$; since the solver
\emph{minimises} $J_i$, hence \emph{maximises} $-J_i$ on that ball,
\[
   -J_i\bigl(\mathbf{o},u^{\mathrm{opt}}(\mathbf{o})\bigr)
   \;\ge\;
   -J_i\bigl(\mathbf{o},a(\mathbf{o})+r\mathbf d\bigr)
   \;\ge\;
   -J_i\bigl(\mathbf{o},a(\mathbf{o})\bigr)+r\delta_2.
\]
Because
$\|\nabla_u Q_i^l(\mathbf{o},u)\|_2\le\delta_1$ in the same ball,
\(
   Q_i^l\bigl(\mathbf{o},u^{\mathrm{opt}}(\mathbf{o})\bigr)
   \;\ge\;
   Q_i^l\bigl(\mathbf{o},a(\mathbf{o})\bigr)-r\delta_1.
\)
Combining the two bounds,
\(
c_1 Q_i^l(\mathbf{o},u^{\mathrm{opt}})
 -c_2 J_i(\mathbf{o},u^{\mathrm{opt}})
 \ge
 c_1 Q_i^l(\mathbf{o},a)
 -c_2 J_i(\mathbf{o},a)
 + r\Delta.
\)
Finally, since
$\bigl|Q^*(\mathbf{o},u)-[c_1Q_i^l(\mathbf{o},u)-c_2J_i(\mathbf{o},u)]\bigr|
          \le\varepsilon$
throughout the ball, we obtain
\[
   Q^*(\mathbf{o},u^{\mathrm{opt}}) + \varepsilon
   \;\ge\;
   c_1 Q_i^l(\mathbf{o},u^{\mathrm{opt}})
      - c_2 J_i(\mathbf{o},u^{\mathrm{opt}})
   \;\ge\;
   c_1 Q_i^l(\mathbf{o},a)-c_2 J_i(\mathbf{o},a)+r\Delta
   \;\ge\;
   Q^*(\mathbf{o},a)-\varepsilon+r\Delta.
\]
Rearranging yields the stated inequality.
\end{proof}

\section{Evaluation Scenarios in Detail}
\label{appendix:experiment_explanations}
 \headline{Experiment: Narrow Corridor.} In this experiment, agents initiated at random locations in a narrow corridor seek to either get to the blue or green region. Agents receive a reward of $-10$ for bumping into each other or the corridor boundaries, $1$ for every time step they spend in the correct region, and a small reward whenever they take a step that brings them closer to this region.

Our handcrafted controller seeks to prevent collisions while sending agents to their target region.  To avoid collisions, the ego agent's controller introduces a CBF defined for each agent $j$: $h_j(\mathbf{p}_{\mathrm{ego}}, \mathbf{p}_j) = k(\|\mathbf{p}_{\mathrm{ego}} - \mathbf{p}_j\|_2^2 - d_{\min}^2)$,
where $d_{\min} > 0$ is a safe distance threshold, $\mathbf{p}$ denotes position, and $k$ is a tuneable constant. This function $h_j$ is positive if the ego agent is at a distance greater than $d_{\min}$ from agent $j$, and negative if too close. The collision avoidance constraints use $h_j$ to ensure that, by properly choosing $\mathbf{u}$, the ego agent moves in a direction that maintains or increases this safety margin, thereby preventing collisions.  At each time step, the ego agent solves (omitting some low-level technical details):
\begin{equation}
\begin{aligned}
\label{eq:narrowcorridorcontroller}
\max_{\mathbf{u}, s} & \quad  d_i\cdot u_y \\
\text{s.t.} \quad
& \forall j, 2\bigl((x_{\mathrm{ego}} - x_j)u_x + (y_{\mathrm{ego}} - y_j)u_y\bigr) + h_j(\mathbf{p}_{\mathrm{ego}}, \mathbf{p}_j) \geq 0 \\
& (\mathbf{p}_{\mathrm{ego}} + \mathbf{u}) \in \mathcal{B}, \|\mathbf{u}\|_2 \leq M \\
\end{aligned}
\end{equation}

Here, $\mathcal{B} = [-X_{\max},X_{\max}] \times [-Y_{\max},Y_{\max}]$ denotes the boundaries of the environment; $d \in \{-1, 1\}$ indicates the target direction of the agent; $\mathbf{u}$ is a decision variable denoting the agent's velocity control input; the first constraint handles agent-agent collisions through the control barrier function; the second constraint prevents agents from going out of bounds; and the third constraint is a velocity limit. To use ReCoDe in this controller, all we need to do is introduce a slack decision variable $s \geq 0$ and a quadratic constraint on top of existing constraints, as shown in \eqref{eq:final_optimization_problem}. Both the handcrafted controller and the ReCoDe modification are efficiently solvable \cite{lobo1998applications}.

Navigating this scenario demands sophisticated agent coordination, as robots must not only avoid collisions but also work in concert to prevent gridlock. While traditional handcrafted controllers can manage basic collision avoidance, they frequently succumb to deadlocks due to their inability to facilitate inter-agent coordination, as illustrated in Figure \ref{fig:experiment scenarios and results}. These limitations expose a fundamental weakness in conventional control approaches that rely on constrained optimization: without a mechanism for coordination, agents cannot make cooperative decisions like yielding to resolve impasses. In contrast, our ReCoDe framework overcomes these challenges.


 \headline{Experiment: Connectivity.} In this experiment, agents in a narrow corridor must move to the green region, while never breaking communication links: every pair of agents needs to always stay within each others' communication range. We introduce static obstacles into the environment that the agents must learn to bypass while maintaining this connectivity requirement. We prevent agent actions that would break this connectivity and give negative rewards to actions that attempt this. Reward structure is otherwise the same as the previous experiment: a positive reward for reaching the green region, and a negative reward for colliding.

Our handcrafted controller is based on \eqref{eq:narrowcorridorcontroller}, with the main difference being the introduction of a quadratic constraint on distance: $\|\mathbf{p}_{\mathrm{ego}} - \mathbf{p}_j\|_2^2 \leq  d_{\min}^2$. 

This problem requires coordination between the agents: an agent greedily attempting to move to the green region might result in a deadlock, as it leaves too little leeway for other agents to bypass obstacles--see Figure \ref{fig:experiment scenarios and results}. While our handcrafted controller struggles in such situations, ReCoDe handles them successfully (see Table \ref{fig:experiment scenarios and results}).

 \headline{Experiment: Waypoint Navigation.} In this experiment, large agents in a small environment must navigate to their respective goal points (Figure \ref{fig:experiment scenarios and results}). The agents and goal points are initiated randomly. At each time step, agents receive a reward proportional to $d_{prev} - d_{current}$ (their previous distance to their goal minus their current distance), as well as a large, discrete reward when they get sufficiently close. A penalty reward of $-10$ is given when agents collide. The scenario is similar to Narrow Corridor in that agents must navigate to a goal point, but unlike Narrow Corridor, agents have a greater diversity of possible goal points and strategies, enabling us to test whether ReCoDe can explore diverse strategies. Our handcrafted controller is based on \eqref{eq:narrowcorridorcontroller}, with only the objective function changed to minimize an agent's distance to its respective goal point.

\begin{figure}
    \centering
    \includegraphics[width=0.5\linewidth]{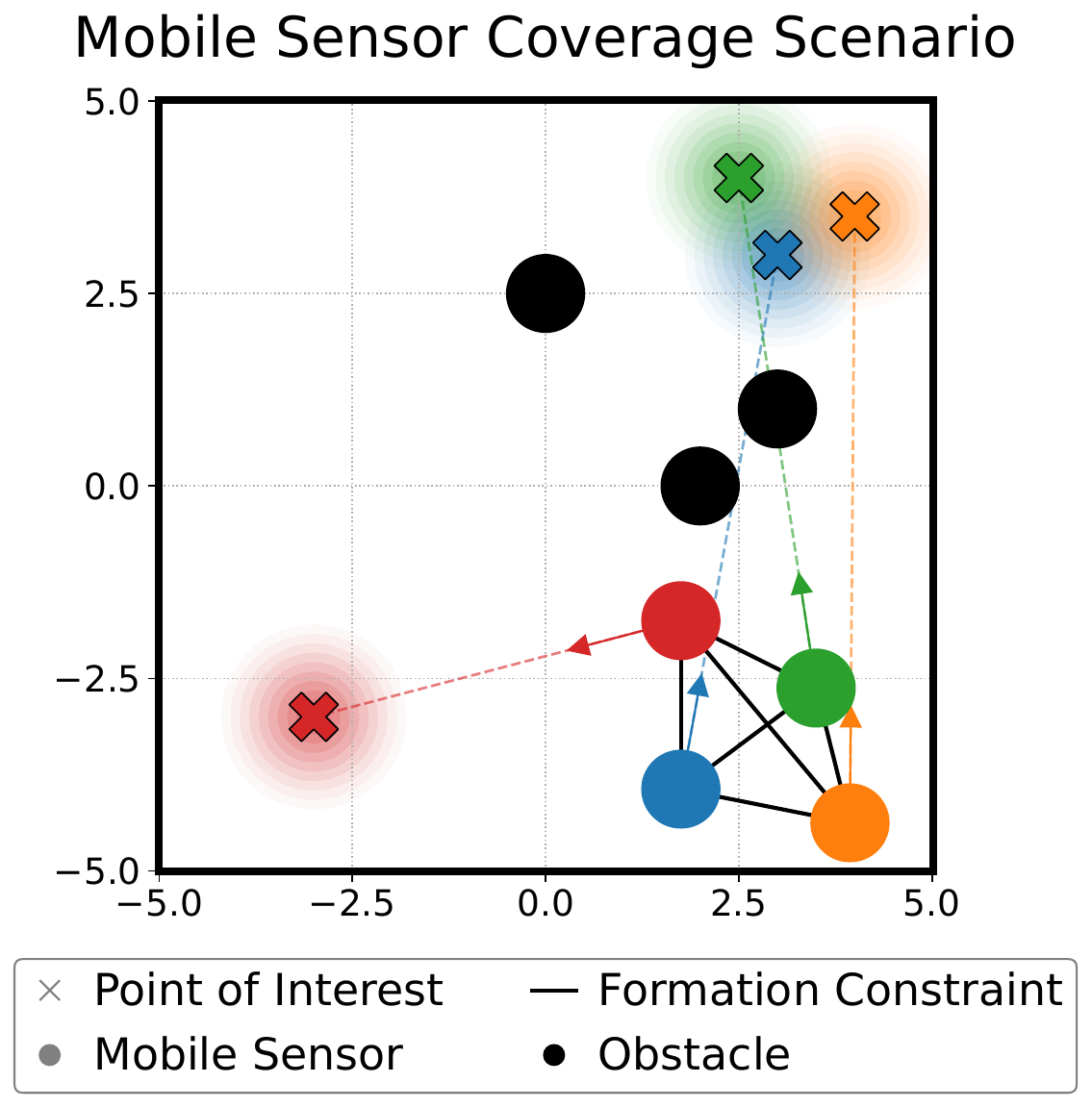}
    \caption{Visual depiction of the \textit{Sensor Coverage} experiment. Sensors (the Os) increase the reward by approaching their color-matched Point of Interest (the Xs) but are constrained by the need to maintain formation with other
sensors.}
    \label{fig:sensorcoverage}
\end{figure}

\headline{Experiment: Sensor Coverage.} 
 \textit{Sensor Coverage} is a multi-objective task that poses a consensus challenge alongside a navigation challenge. In it, decentralized mobile sensors monitor different phenomena (e.g. wildlife or pollution) distributed across different locations--see Figure \ref{fig:sensorcoverage}. The sensors are placed in an environment with obstacles. Each sensor aims to maximize accuracy by getting as close as possible to its assigned sensing target (which is unique to it), but the same formation constraints as in the \textit{Connectivity} scenario apply: sensors must maintain all communication links by staying together. Due to this formation constraint, the sensors must choose which targets to prioritize and decide on  the best location for overall coverage, which might, e.g., be directly on top of some assigned targets, or somewhere near the center of the targets' convex hull, not reaching any target but weakly covering all targets, thus maximizing cumulative reward. The agents  must also all coordinate about how to move in formation to their desired location.
 
 At every time-step the total reward $r_i$ for sensor~$i$ is $
r_i = r^{\mathrm{prox}}_i + r^{\mathrm{safety}}_i
$. Here,  $r^{\mathrm{prox}}_i = e^{-\lambda_{prox} \bigl\lVert\mathbf{p}_i-\mathbf{g}_i\bigr\rVert^2}$, where $\lambda_{prox}$ is a hyperparameter,  $\mathbf{p}_i$ is the sensor’s position and $\mathbf{g}_i$ the centre of its sensing target.  $ r^{\mathrm{safety}}_i$ denotes a reward penalty for unsafe behavior--collisions, going out of bounds, and breaking formation constraints, as in the \textit{Connectivity} scenario.

The handcrafted QP is identical to~\eqref{eq:narrowcorridorcontroller} \emph{except} that  
the objective minimizes the distance-to-goal $\lVert\mathbf{p}_i+\mathbf{u}_i-\mathbf{g}_i\rVert_2^2$. This kind of objective is suboptimal since it pulls agents in different directions (due to having different goals), but they must stay connected, hence the entire cloud of agents can get stuck in a deadlock. However, it is the best we could find for a quadratic constrained optimization program, and it performs on par with other baselines except ReCoDe--see Table \ref{fig:experiment scenarios and results}.

This is a multi-objective problem requiring decentralized consensus where the sensors must determine \textit{where} to move and \textit{how} to get there effectively. Discovering, and agreeing, on a solution to pursue along the Pareto frontier of optimal trajectories and final locations is hard. Shielding and Pure MARL have a difficulty optimizing this delicate trade-off, and the fixed QP stalls when two attractive targets pull the team in opposite directions. ReCoDe’s learned constraints let neighboring sensors negotiate their trajectory, while locally relying on the default QP's (imperfect) objective function to further improve performance--thus achieving superior reward on this task. 

\section{Additional Details on Robot Demonstrations}
\label{appendix:differentconstraintforms}

The platform we used is the Cambridge Robomaster \cite{blumenkamp2024cambridge}. We used the Freyja library \cite{shankar2021freyja} for executing velocity commands. Our simulator was implemented using the \texttt{robomaster} branch of the BenchMARL repository \cite{bettini2024benchmarl} as a base. We implemented our Narrow Corridor scenario in this repository, and manually calibrated the robot controller parameters and arena dimensions to attain sim2real. The default QP used for the robots is the one defined for the Narrow Corridor scenario in Appendix \ref{appendix:experiment_explanations}, lightly calibrated to attain better sim2real performance. We used motion capture for localization, and messages were only passed between robots within each others' communication range (\SI{1.5}{\meter}). However, robots were only made aware of  *relative spatial information* about neighboring robots and their own coordinates (relative velocity and position).

\textbf{What if we learn linear constraints instead of quadratic constraints?} We experimented with having robots learn a single linear constraint of the form  $\mathbf{a}_i^\top \mathbf{u}_i(t) \leq b_i$, where $\mathbf{a}_i \in \mathbb{R}^m$ and $b_i \in \mathbb{R}$ are parameters that the agent can adjust. In Appendix \ref{appendix:whynotlinearconstraints} we explain why these kind of constraints are generally inefficient. However, because the default QP we used for the Narrow Corridor scenario already introduces other constraints (control barrier function and boundary constraints to prevent collisions), linear constraints become more expressive, as they can \textit{intersect} these other constraints and bound various kinds of volumes. Thus, we speculated that they will perform well in this scenario.

We found that using the linear constraint, although it is less principled, neither decreases nor increases average reward or robomaster performance. However, it seems to enforce different kinds of behaviors: while the quadratic constraint sometimes drove robots to attain \textit{partial} solutions of the Narrow Corridor environment (by having only some of them reach their goal), the linear constraint-trained robots seemed to adapt an ``all or nothing'' strategy, where either all robots reach their goal, or none of them do. It is interesting that both strategies led to the same average reward despite these different behavioral modes. In future work, we are interested in exploring the effects of different kinds of constraints on the agents'  learned policy.

\end{document}